\documentclass[letterpaper, 10 pt, conference]{ieeeconf}  

\IEEEoverridecommandlockouts                              

\makeatletter
\let\NAT@parse\undefined
\makeatother

\usepackage[numbers]{natbib}
\usepackage{multicol}
\usepackage{hyperref}
\hypersetup{bookmarksopen,bookmarksnumbered,
pdfpagemode=UseOutlines,
colorlinks=true,
linkcolor=blue,
anchorcolor=blue,
citecolor=blue,
filecolor=blue,
menucolor=blue,
urlcolor=blue,
breaklinks=true
}
\let\proof\relax

\usepackage{amsmath,amsthm,amsfonts,mathtools}
\usepackage{upgreek}
\usepackage{tikz}
\usepackage{braids}
\usepackage{subcaption}

\renewcommand{\dotsb}{%
  \mathinner{\cdotp\mkern-3mu\cdotp\mkern-3mu\cdotp}%
}

\newcommand{\mydots}{\scalebox{0.3}{$\dotsb$}}
\usepackage{amssymb}  
\usepackage{pifont}
\usepackage{cite}
\usepackage{epstopdf}
\usepackage{float}
\newfloat{algorithm}{t}{lop}
\usepackage{stfloats}
\usepackage{mathtools}
\usepackage{algorithm}
\usepackage{algorithmicx}
\usepackage[noend]{algpseudocode}
\usepackage{booktabs}
\usepackage{siunitx}
\DeclareSIUnit \fps {fps}

\let\labelindent\relax
\usepackage{enumitem}

\theoremstyle{remark}
\newtheorem{theorem}{Theorem}[section]

\usepackage{calrsfs}
\DeclareMathAlphabet{\pazocal}{OMS}{zplm}{m}{n}
\DeclareMathAlphabet\mathbfcal{OMS}{zplm}{b}{n}

\usepackage{balance}

\newcommand{\figref}[1]{Fig.~\ref{#1}}

\usepackage{bm}

\usepackage{url}

\newcommand{\xxnote}[3]{}
\ifx\hidenotes\undefined
  \usepackage{color}
  \renewcommand{\xxnote}[3]{\color{#2}{#1: #3}}
\fi

\title{\LARGE \bf Analyzing Multiagent Interactions in Traffic Scenes\\ via Topological Braids}

\author{Christoforos Mavrogiannis$^{1}$, Jonathan DeCastro$^{2}$, and Siddhartha S. Srinivasa$^{1}$
\thanks{This work was (partially) funded by the National Science Foundation IIS (\#2007011), National Science Foundation DMS (\#1839371), the Office of Naval Research, US Army Research Laboratory CCDC, Amazon, and Honda Research Institute USA.}
\thanks{$^{1}$C. Mavrogiannis and S. S. Srinivasa are with the Paul G. Allen School of Computer Science \& Engineering, University of Washington, Seattle, WA 98102, USA.
        {E-mail: \tt\small \{cmavro, siddh\}@cs.washington.edu}.}%
\thanks{$^{2}$J. A. DeCastro is with the Toyota Research Institute, Cambridge, MA 02139, USA.
        {E-mail: \tt\small jonathan.decastro@tri.global}.}%
}

\begin{document}

\maketitle
\thispagestyle{empty}
\pagestyle{empty}

\begin{abstract}
We focus on the problem of analyzing multiagent interactions in traffic domains. Understanding the space of behavior of real-world traffic may offer significant advantages for algorithmic design, data-driven methodologies, and benchmarking. However, the high dimensionality of the space and the stochasticity of human behavior may hinder the identification of important interaction patterns. Our key insight is that traffic environments feature significant geometric and temporal structure, leading to highly organized collective behaviors, often drawn from a small set of dominant modes. In this work, we propose a representation based on the formalism of topological braids that can summarize arbitrarily complex multiagent behavior into a compact object of dual geometric and symbolic nature, capturing critical events of interaction. This representation allows us to formally enumerate the space of outcomes in a traffic scene and characterize their complexity. We illustrate the value of the proposed representation in summarizing critical aspects of real-world traffic behavior through a case study on recent driving datasets. We show that despite the density of real-world traffic, observed behavior tends to follow highly organized patterns of low interaction. Our framework may be a valuable tool for evaluating the richness of driving datasets, but also for synthetically designing balanced training datasets or benchmarks.

\end{abstract}

\section{Introduction}\label{sec:introduction}



Traffic scenes pose unique challenges for prediction and planning~\citep{Sadigh-RSS-16,gadepally_framework_2017,Roh2020Multimodal,DeCastroLAP20,bgap} due to their high dimensionality, the complexity of modeling human behavior, and the performance standards motivated by the safety-critical nature of the domain. Despite these complications, real-world traffic scenes often feature significant structure. Vehicles follow designated lanes, and traffic is regulated through traffic signals and signs or coordinated via turn indicators. Driver behavior can often be modeled as rational, characterized by risk aversion and efficiency-seeking objectives. Recent work has leveraged these observations in the design of data-driven models for behavior prediction and planning~\citep{tian19,Bouton17,Roh2020Multimodal,hsu18,trajectron}. To measure and transfer their performance to the real world, such models require large, balanced datasets, containing a diversity of behavior that is representative of real-world traffic. This requires an understanding of the space of behavior that typically unfolds in different scenes. While there exist tools for characterizing various aspects of behavior~\citep{tolstaya2021identifying,Liebenwein20,DeCastroLAP20,bgap}, summarizing multiagent interaction in an intuitive and formal way is hindered by the high dimensionality and the stochasticity of human behavior.




\begin{figure}
    \centering
    \includegraphics[width = \linewidth]{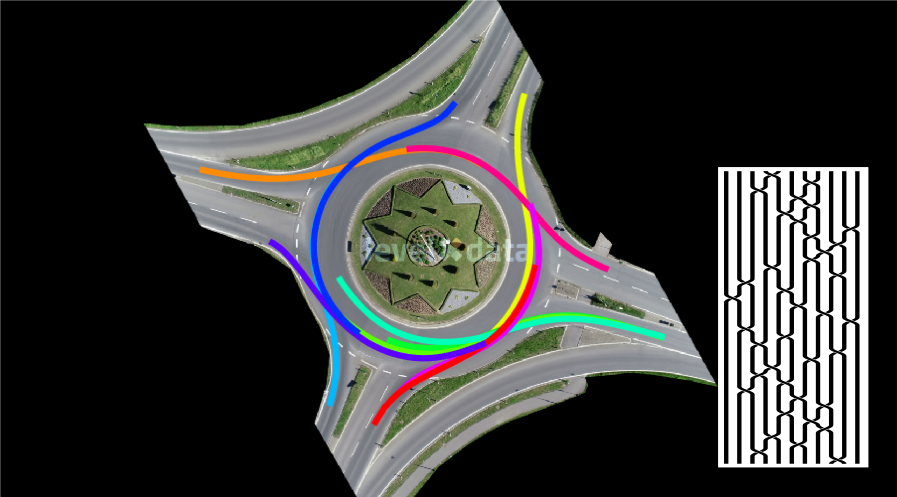}
    \caption{This work proposes a formal framework for the characterization of multiagent behavior in driving domains. Complex multiagent interactions encountered in real-world driving domains such as a roundabout can be compactly represented as topological braids (right).}
    \label{fig:figure1}
\end{figure}

Our key insight is that the behavior encountered in traffic scenes often exhibits prevalent topological structure: common events like overtaking, merging or traversing a roundabout (\figref{fig:figure1}) give rise to \emph{interactions} of distinct topological signatures~\citep{Berger2001hamiltonians}. In this work, we abstract multiagent behavior in a traffic scene into a topological braid~\citep{artin}, a compact and interpretable topological object with symbolic and geometric descriptions. Building on past work on the use of braids for multiagent navigation~\citep{mavrogiannis_ijrr}, we make the following contributions. First, we adapt the representation of~\citet{mavrogiannis_ijrr} to structured domains like driving environments through a rigorous mathematical presentation. We then study its computational properties, and discuss how a measure of complexity based on braids~\citep{Dynnikov2007} may capture the interactivity of a traffic scene. We show that our framework is applicable to complex scenes through a case study on real-world intersections and roundabouts~\citep{inDdataset,rounDdataset}. We cluster the behaviors exhibited in these scenes into braids, and characterize their complexity. We find that in the majority of scenes, a few simple braids dominate, indicating a low degree of interaction despite the high traffic density. Our methodology can be valuable for the analysis and design of road networks, the design and benchmarking of data-driven frameworks for prediction and planning, the evaluation and generation of driving datasets.

\section{Related Work}\label{sec:related-work}


Recent work on behavior prediction and decision making for autonomous driving applications has leveraged discrete, semantic representations of traffic behavior. For instance, \citet{wang_understanding_2018} classify discrete driving styles using a variant of hidden Markov models (HMM). \citet{gadepally_framework_2017} also use HMM to estimate long-term driver behaviors from a sequence of discrete decisions. Others, such as \citet{konidaris_skills_2018} and \citet{shalev-shwartz_safe_2016}, propose using learned symbolic representations for high-level planning and collision avoidance, via a hierarchical options model. Similarly, \citet{TOPSinteractive} learn a latent representation of interactions. While these works uncover discrete representations of driving behavior, they either require large datasets to learn discrete modes or specify them manually without harnessing the rich geometric structure of the environment.

Another body of work has focused on tools for testing and validation in realistic settings, leveraging a semantic-level understanding of interactions. \citet{tian19} model traffic at unsignalized intersections using tools from game theory and propose a verification testbed for navigation algorithms. \citet{Liebenwein20} propose a framework for safety verification of driving controllers based on compositional and contract-based principles. 
\citet{hsu18} investigate how velocity signals generated by Markov decision processes affect interaction dynamics at intersections.   
\citet{DeCastroLAP20} construct a representation of multi-vehicle interaction outcomes based on latent parameters using a generative model.
\citet{tolstaya2021identifying} propose an \emph{Interactivity} score that enables the identification of interesting interactive scenarios for training generative models. Our work is similar in spirit and complementary to this latter line of work. We also approach a notion of interactivity between agents. However, instead of investigating statistical properties of distributions, we focus on the aspect of the representation, through the introduction of tools from low-dimensional topology.


Recently, roboticists have been increasingly making use of topological representations to model the rich structure of real-world domains. These include knitting~\citep{knitting-icra}, untangling~\citep{untangling} or knot planning~\citep{bohg-knot-planning}, aircraft conflict resolution~\citep{hu-braids2000} or multiagent navigation~\citep{diazmercado}. Some works leverage insights from homotopy theory~\citep{dynamic-channel,Bhattacharya18}, persistent homology~\citep{pokorny16} and fiber bundles~\citep{orthey2020multilevel}. Some other works make use of topological abstractions such as invariants~\citep{MavrogiannisEtal_HRI_2018,mavrogiannis-hamiltonians,Roh2020Multimodal} and braids~\citep{MavBluKne_IROS2017,mavrogiannis_ijrr} as representatives of multiagent motion primitives for prediction and planning. In this paper, we are following up on this latter body of work work by employing topological braids as an abstraction of traffic behavior. While past work considered simplified simulation domains~\citep{MavBluKne_IROS2017,mavrogiannis_ijrr}, in this paper, we adapt the braid presentation to accommodate rich traffic environments such as real-world intersections or roundabouts. To the best of our knowledge this paper is the first to investigate the applicability of braids as primitives for multiagent behavior in realistic real-world environments. 

\section{Abstracting Driving Interactions as Topological Braids}\label{sec:multimodality}

We introduce a representation based on topological braids~\citep{artin}, that captures critical interaction events in street environments (e.g., overtaking, merging, crossing). 
This representation describes such interactions as sequences of symbols describing topological relationships between agents; any possible interaction manifests as a unique symbolic representation of their trajectories.
Our representation adapts the presentation of \citet{mavrogiannis_ijrr} to real-world traffic domains through theoretical developments. 



\subsection{Domain}

Consider a structured domain $\pazocal{Q}\subseteq\mathbb{R}^2$ where $n > 1$ agents are navigating from time $t=0$ to a finite final time $t_{\infty}$. Define by $q_{i}\in\pazocal{Q}$ the position of agent $i\in \pazocal{N} = \{1,\dots, n\}$ with respect to a fixed reference frame. Agent $i$ follows a trajectory $\xi_i:[0,t_{\infty}]\to\pazocal{Q}$. Collectively, agents follow a system trajectory $\Xi = (\xi_{1},\dots, \xi_{n})$. This trajectory is a detailed representation of the collective strategy that agents followed to avoid each other while following their paths. Their strategy can be summarized as a set of discrete relationships, such as the passing sides or crossing order of agents. These relationships are formed as a result of the geometric structure of the environment, traffic regulations, and agents' policies. In this paper, we show that such relationships feature topological properties that can be succinctly captured by the representation of topological braids~\citep{artin}.





\begin{figure}
\centering
    \begin{subfigure}{.48\linewidth}
\scalebox{.6}{
  \begin{tabular}{c c c}  
    \begin{minipage}[b]{0.3\linewidth}
    \centering
    \begin{tikzpicture}
      \braid[number of strands=5, line width=2pt, width = 10pt, height = 55pt,  style strands={1}{black},
 style strands={2}{red},
 style strands={3}{black},  style strands={4}{black}, style strands={5}{black}]
      a_1^{-1} a;
      \node[font=\Huge] at (1.6,-1.2) { \(\mydots\) };
    \end{tikzpicture}\\
    \small{$\sigma_{1}$}
    \end{minipage}    
        \begin{minipage}[b]{0.1\linewidth}
    \centering
    \raisebox{3.9em}{
          \begin{tikzpicture}
      \node[font=\large] { \(,\) };
    \end{tikzpicture}}\\
    \end{minipage}
    \begin{minipage}[b]{0.3\linewidth}
    \centering
          \begin{tikzpicture}
      \braid[number of strands=5, line width=2pt, width = 10pt, height = 55pt,  style strands={1}{black},
 style strands={2}{black},
 style strands={3}{red},  style strands={4}{black},  style strands={5}{black}]
      a_2^{-1} a;
      \node[font=\Huge] at (1.6,-1.2) { \(\mydots\) };
    \end{tikzpicture}\\
          \small{$\sigma_{2}$}
    \end{minipage}    
    \begin{minipage}[b]{0.28\linewidth}
    \centering
    \raisebox{3.9em}{
          \begin{tikzpicture}
      \node[font=\large] { \(,\dots,\) };
    \end{tikzpicture}}\\
    \end{minipage}
    
    \begin{minipage}[b]{0.3\linewidth}
    \centering
          \begin{tikzpicture}
      \braid[number of strands=5, line width=2pt, width = 10pt,  height = 55pt,  style strands={1}{black},
 style strands={2}{black},
 style strands={3}{black},  style strands={4}{black},  style strands={5}{red}]
      a_4^{-1} a;
      \node[font=\Huge] at (.9,-1.2) { \(\mydots\) };
    \end{tikzpicture}\\
          \small{$\sigma_{n-1}$}
    \end{minipage}\\  
     & &  
  \end{tabular}
  }
\caption{\small{Generators of $B_{n}$.}\label{fig:BnGenerators}}
\end{subfigure}
    \begin{subfigure}{.48\linewidth}
        \centering
\scalebox{.6}{
  \begin{tabular}{c c c}  
    \begin{minipage}[b]{0.3\linewidth}
    \centering
    \begin{tikzpicture}
      \braid[number of strands=5, line width=2pt, width = 10pt, height = 55pt, style strands={1}{black},
 style strands={2}{red},
 style strands={3}{blue},  style strands={4}{black},  style strands={5}{black}]
      a_1^{-1} a;
      \node[font=\Huge] at (1.6,-1.2) { \(\mydots\) };
    \end{tikzpicture}\\
    \small{$\sigma_{1}$}
    \end{minipage}    
    
        \begin{minipage}[b]{0.1\linewidth}
    \centering
    \raisebox{3.9em}{
          \begin{tikzpicture}
      \node[font=\large] { \(\cdot\) };
    \end{tikzpicture}}\\
    \end{minipage}
        
    \begin{minipage}[b]{0.3\linewidth}
    \centering
          \begin{tikzpicture}
      \braid[number of strands=5, line width=2pt, width = 10pt, height = 55pt, style strands={1}{black},
 style strands={2}{blue},
 style strands={3}{red},  style strands={4}{black},  style strands={5}{black}]
      a_2 a;
      \node[font=\Huge] at (1.6,-1.2) { \(\mydots\) };
    \end{tikzpicture}\\
          \small{$\sigma_{2}^{-1}$}
    \end{minipage}    
    ~~~~
    \begin{minipage}[b]{0.1\linewidth}
    \centering
    \raisebox{3.9em}{
          \begin{tikzpicture}
      \node[font=\large] { \(=\) };
    \end{tikzpicture}}\\
    \end{minipage}
    ~~~~
    \begin{minipage}[b]{0.3\linewidth}
    \centering
          \begin{tikzpicture}
      \braid[number of strands=5, line width=2pt, width = 10pt,  height = 27.5pt, style strands={1}{black},
 style strands={2}{blue},
 style strands={3}{red},  style strands={4}{black},  style strands={5}{black}]
      a_2 a_1^{-1}  ;
      \node[font=\Huge] at (1.6,-1.2) { \(\mydots\) };
    \end{tikzpicture}\\
          \small{$\sigma_{1}\cdot\sigma_{2}^{-1}$}
    \end{minipage}\\  
     & &  
  \end{tabular}
  }
\caption{\small{Example of composition.}\label{fig:composition}}
\end{subfigure}
\caption{Presentation of the braid group, $B_n$.}
\end{figure}
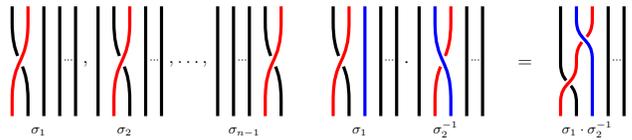

\subsection{Topological Braids\label{sec:braids}}


A braid is a tuple $b_f = (f_{1},\dots, f_{n})$ of functions $f_{i}:I\to\mathbb{R}^2\times I$, $i\in \pazocal{N}$, defined on a domain $I = [0,1]$ and embedded in a Cartesian space $(\hat{x},\hat{y},\hat{t})$. These functions, called \textit{strands}, are monotonically increasing along the $\hat{t}$ direction, satisfying the properties: (a) $f_{i}(0) = (i,0,0)$, and $f_{i}(1) = (p_{f}(i),0,1)$, where $p_{f}: \pazocal{N}\to \pazocal{N}$ is a permutation in the symmetric group $N_{n}$; (b) $f_{i}(t)\neq f_{j}(t)$ $\forall$ $t\in I$, $j\neq i\in \pazocal{N}$. Two braids, $b_{f} = (f_{1},\dots,f_{n})$, $b_{g} = (g_{1},\dots,g_{n})$, can be composed through a \emph{composition} operation (\figref{fig:composition}): their composition, $b_{h} = b_{f}\cdot b_{g}$, is also a braid $b_{h} = (h_{1},\dots, h_{n})$, comprising a set of $n$ curves, defined as:
\begin{equation}
h_{i}(t) = \begin{dcases} f_{i}(2t), & t\in\left[0,\frac{1}{2}\right) \\ g_{j}(2t-1), & t\in\left[\frac{1}{2},1\right]\end{dcases}\mbox{,}
\end{equation}
where $j = p_{f}(i)$. The set of all braids on $n$ strands, along with the composition operation form a group, $B_{n}$, called the Braid group on $n$ strands. Following Artin's presentation \citep{artin}, the braid group $B_{n}$ can be generated from $n-1$ primitive braids $\sigma_{1},...,\sigma_{n-1}$ (see \figref{fig:BnGenerators}), called generators, and their inverses, via composition.


A \textbf{generator} is a braid $\sigma_{i} = (\upsigma_{1},\dots,\upsigma_{n})$, $i\in \pazocal{N}\setminus \{n\}$ for which: (a) $\upsigma_{i}(0) = (1,0,0)$, and $\upsigma_{i}(1) = (p_{i}(i),0,1)$, where $p_{i}: \pazocal{N}\to \pazocal{N}$ is an adjacent transposition swapping $i$ and $i+1$; (b) there exists a unique $t_{c}\in [0,1]$ such that $(\upsigma_{i}(t_{c})-\upsigma_{i+1}(t_{c}))\cdot \hat{x} = 0$ and $(\upsigma_{i}(t_{c})-\upsigma_{i+1}(t_{c}))\cdot \hat{y} >0$. 
%

%

The \textbf{inverse} of $\sigma_{i}$ is the braid $\sigma^{-1}_{i}= (\upsigma^{-1}_{1},\dots,\upsigma^{-1}_{n})$, $i\in \pazocal{N}\backslash n$, for which: (a) $\upsigma^{-1}_{i}(0) = (1,0,0)$, and $\upsigma^{-1}_{i}(1) = (p_{i}(i),0,1)$; (b) there exists a unique $t_{c}\in [0,1]$ such that $(\upsigma^{-1}_{i}(t_{c})-\upsigma^{-1}_{i+1}(t_{c}))\cdot \hat{x} = 0$ and $(\upsigma^{-1}_{i}(t_{c})-\upsigma^{-1}_{i+1}(t_{c}))\cdot \hat{y} < 0$. 

The \textbf{identity} braid $\sigma_{0} = (\upsigma^{0}_{1},\dots, \upsigma^{0}_{n})$ implements no swap, i.e., $p_{0}(k) = k$ for any $k\in \pazocal{N}\setminus \{n\}$, yielding $\upsigma^{0}_{k}(0) = (k,0,0)$, $\upsigma^{0}_{k}(1) = (k,0,1)$ and it holds that $\nexists t_{c}\in [0,1]$ such that $(\upsigma^{0}_{k}(t_{c})-\upsigma^{0}_{k+1}(t_{c}))\cdot \hat{x} = 0$. 

Any braid can be written as a \emph{word}, i.e., a product of generators and their inverses (\figref{fig:composition}), satisfying the \textit{relations}:
\begin{equation}
\begin{split}
&\sigma_{i}\sigma_{j} =\sigma_{j}\sigma_{i},\:|j-i|>1,\\
&\sigma_{i}\sigma_{i+1}\sigma_{i}=\sigma_{i+1}\sigma_{i}\sigma_{i+1},\:\forall\: i\mbox{.}\label{eq:braidrelations}
\end{split}
\end{equation}

\subsection{Transforming Traffic Trajectories into Braids}\label{sec:trajectoriesbraids}




We will convert a system trajectory $\Xi$ into a Cartesian object with the structure of a topological braid through a sequence of operations that retain the topological relationships among agents' trajectories.

We define by $\xi^{x}_{i}:[0,t_{\infty}]\to\mathbb{R}$ and $\xi^{y}_{i}:[0,t_{\infty}]\to\mathbb{R}$ the $x$ and $y$ projections of $\xi_i$. For $t=0$, ranking agents in order of increasing $\xi^{x}_{i}(0)$, $i\in\pazocal{N}$ value defines a starting permutation $p_{s}:\pazocal{N}\to \pazocal{N}$, where $p_{s}(i)$ denotes the order of agent $i$. For $t=t_{\infty}$, ranking agents in order of increasing $\xi^{x}_{i}(t_{\infty})$ value defines a final permutation $p_{d}:\pazocal{N}\to \pazocal{N}$, where $p_{d}(p_{s}(i))$ denotes the final ranking of agent $i$. Thus the execution in $\Xi$ can be abstracted into a transition from $p_s$ to $p_d$.

\begin{figure}
    \centering
    \begin{subfigure}{0.48\linewidth}
        \centering
        \includegraphics[trim = {0cm 0cm 0cm 0cm}, clip, width = \linewidth]{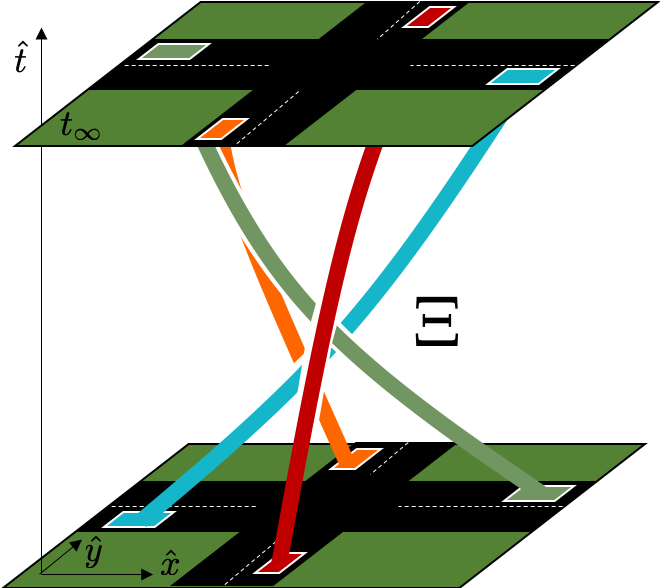}
        \caption{Trajectories of four agents as they navigate an intersection, plotted in spacetime.\label{fig:intersection-perspective}}
    \end{subfigure}
    ~
    \begin{subfigure}{0.48\linewidth}
        \centering
        \includegraphics[trim = {0cm 0cm 0cm 0cm}, clip, width = \linewidth]{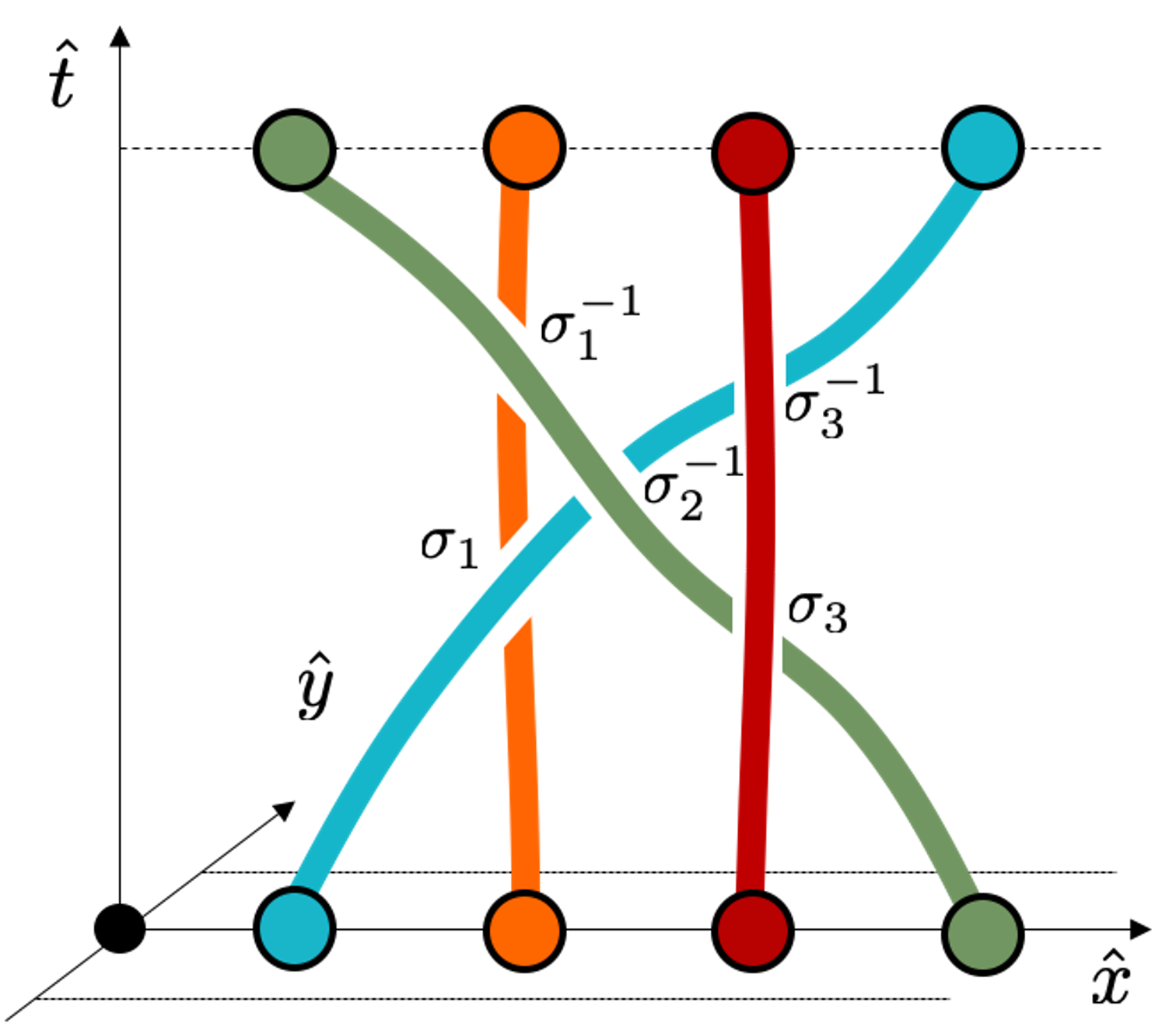}
        \caption{Braid $\sigma^{\phantom{-}}_{3}\sigma_{1}^{\phantom{-}}\sigma_{2}^{-1}\sigma^{-1}_{3}\sigma_{1}^{-1}$ capturing the topological entanglement of agents' trajectories.\label{fig:intersection-braid}}
    \end{subfigure}    
    \caption{Transition from Cartesian trajectories (\subref{fig:intersection-perspective}) to topological braids (\subref{fig:intersection-braid}) via eq.~\eqref{eq:xiplus} assuming a $x$-$t$ projection.}
    \label{fig:braidtransitionexamples}
\end{figure}


We denote by $\uptau:I\to [0, t_{\infty}]$ a time normalization function, uniformly mapping $I=[0,1]$ to the execution time in the range $[0,t_{\infty}]$. We then define the trajectory bounds as $x_{min} = \min_{i,t} \xi^{x}_{i}(t)$, $x_{max} = \max_{i,t} \xi^{x}_{i}(t)$, and $y_{min} = \min_{i,t} \xi^{y}_{i}(t)$, $y_{max} = \max_{i,t} \xi^{y}_{i}(t)$. Assuming $x_{max}\neq x_{min}$, $y_{max}\neq y_{min}$, we define the ratio functions:
\begin{equation}
    r_{i}^{x}(t) = \frac{\xi^{x}_{i}(t)-x_{min}}{x_{max} - x_{min}}\mbox{,}
\qquad
    r_{i}^{y}(t) = \frac{\xi^{y}_{i}(t)-y_{min}}{y_{max} - y_{min}}\mbox{.}
\end{equation}
Finally, we define a set of functions $\left(f_{1},\dots,f_{n}\right)$, with $f_{j}:I\to\mathbb{R}^2\times I$, $j\in \pazocal{N}$, such that:
\begin{equation}
\begin{split}
& f_{j}(a) =\begin{dcases}
\left(j,0,0\right), & a = 0\\
(f_j^x(a), f_j^y(a), a), & a \in \left(0,1\right)\\
\left(p_{d}(j),0,1\right), & a = 1\\
\end{dcases}
\end{split}
\mbox{,}
\label{eq:xiplus}
\end{equation}
where 
\begin{equation}
\begin{split}
f_j^x &=  1+ r_{j}^{x}(\uptau(a))(n-1),\\ f_j^y &= -1 + 2r_{j}^{y}(\uptau(a))
\mbox{,}
\end{split}
\end{equation}
and $j = p_{s}(i)$, $i\in\pazocal{N}$. For $a\in (0,1)$, the expressions of \eqref{eq:xiplus} scale $x$-coordinates to lie within $[1,n-1]$ and the $y$-coordinates to lie within $[-1,1]$ in a way that preserves topological relationships among trajectories. The set of functions $\left(f_1,\dots, f_n\right)$ is a topological \emph{braid} $\beta$ following the definition of Sec.~\ref{sec:braids}. The braid $\beta$ is topologically equivalent --ambient-isotopic \citep{murasugi1999study}---to the system trajectory $\Xi$. 

\begin{figure}
\centering
\begin{subfigure}{.31\linewidth}
\centering
\includegraphics[width = \linewidth]{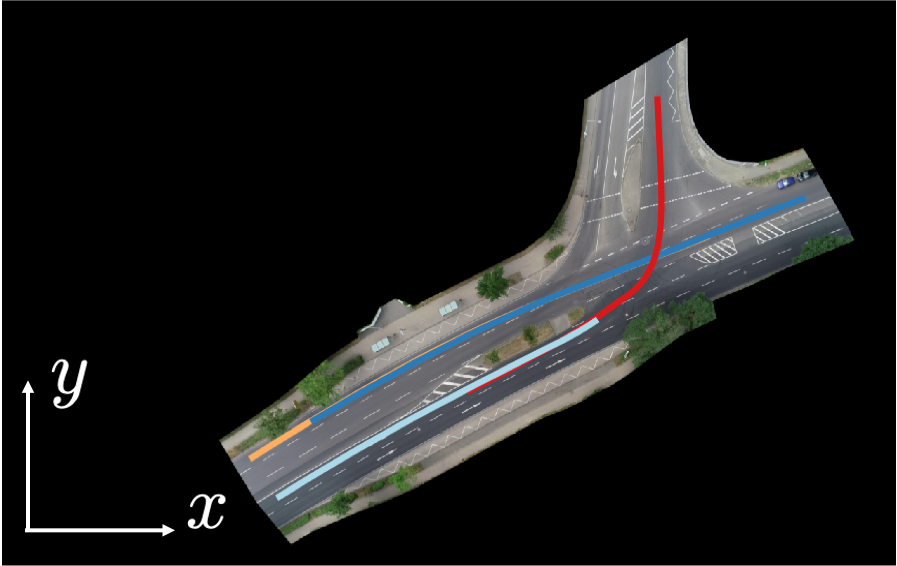}
\caption{Top view.\label{fig:transition-scene}}
\end{subfigure}
~
\begin{subfigure}{.31\linewidth}
\centering
\includegraphics[width = \linewidth]{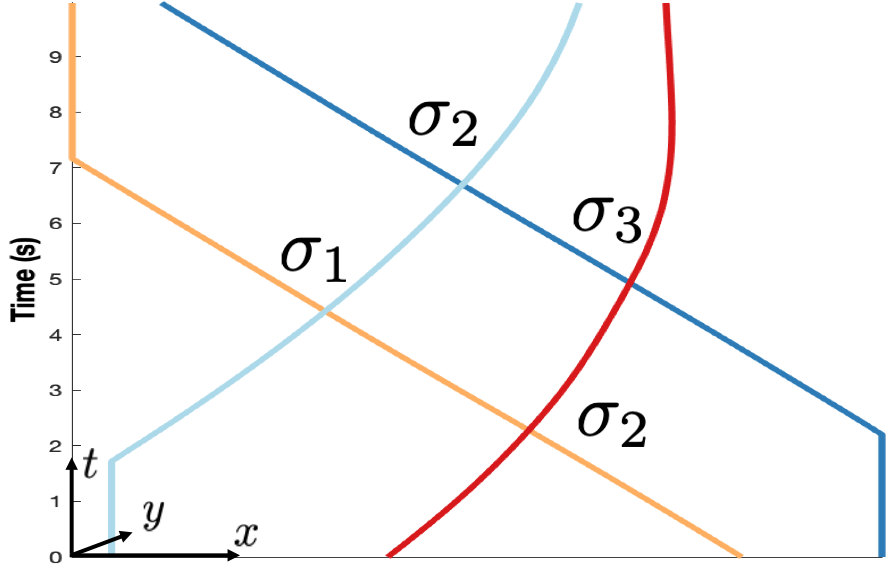}
\caption{Side projection.\label{fig:transition-projection}}
\end{subfigure}
~
\begin{subfigure}{.3\linewidth}
\centering
\includegraphics[width = \linewidth]{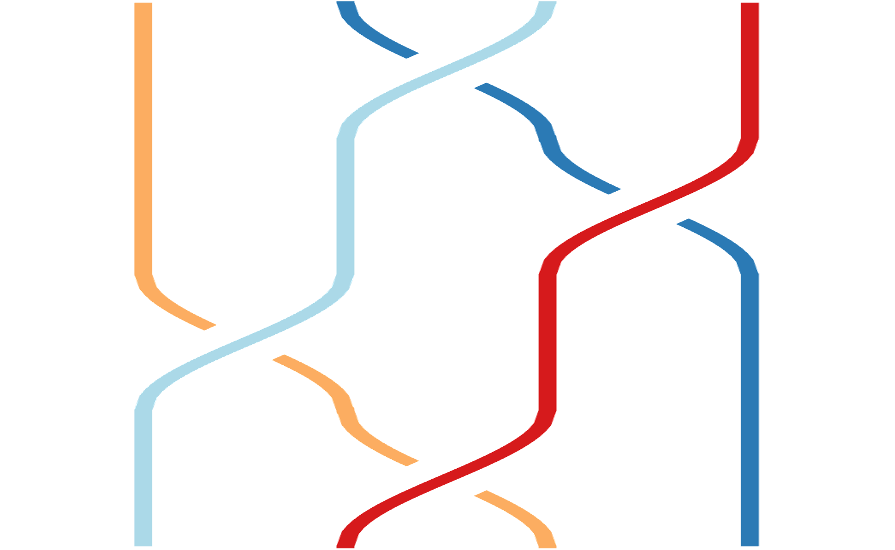}
\caption{Extracted braid.\label{fig:transition-braid}}
\end{subfigure}
\caption{Transitioning from a real-world episode to a braid. The trajectories of (\subref{fig:transition-scene}) are first projected on the plane $x$-$t$ (\subref{fig:transition-projection}) and then the braid $\sigma_{2}\sigma_{1}\sigma_{3}\sigma_{2}$ is reconstructed (\subref{fig:transition-braid}).\label{fig:transition}}
\end{figure}



\subsection{Braids as Modes of Traffic Coordination}

The transformation of Sec.~\ref{sec:trajectoriesbraids} enables summarization of a traffic episode into a braid capturing multiagent collision-avoidance relationships. This braid can be written as a word, similarly to how \citet{Thiffeault2010} converted particle motion in a fluid to a braid (\figref{fig:braidtransitionexamples}): a) we label any trajectory crossings that emerge within the $x$-$t$ projection as braid generators by identifying \emph{under} or \emph{over} crossings (\figref{fig:intersection-braid}); b) we arrange these generators in temporal order into a \emph{braid word}. Note that alternative reference frames can be employed; we selected the $\hat{x}$-$\hat{t}$ plane projection for convenience.



In~\figref{fig:braidtransitionexamples}, four agents cross an intersection. The braid $\sigma^{\phantom{-}}_{3}\sigma_{1}^{\phantom{-}}\sigma_{2}^{-1}\sigma^{-1}_{3}\sigma_{1}^{-1}\in B_{4}$ is a description of how agents coordinated to avoid each other. The group $B_{4}$ contains all ways in which these four agents could possibly avoid each other. In a scene with $n$ agents, a braid represents a \emph{mode} of coordination from the set of possible modes in $B_n$.

\subsection{Computational Properties of the Representation\label{sec:properties}}

To highlight the possible computational benefits arising from the summarization of traffic episodes into braids, we study the runtime of enumerating modes of coordination as topological braids in comparison to enumerating Cartesian trajectories. Consider a traffic episode of $H$ timesteps, involving $n$ agents. Each agent has $\pazocal{T}$ options of routes to follow and $\pazocal{U}$ actions to take at every timestep. We assume that there is at most one agent per lane, i.e., $n\leq\pazocal{T}$. The horizon of the execution is long, and thus $n \ll H$. Further, $\pazocal{U}$ is a realistically rich space of controls, and thus $n \ll\pazocal{U}$, $H \ll \pazocal{U}$ and $\pazocal{T}  \ll \pazocal{U}$. Finally, we assume that agents are goal-driven for the horizon of each episode, and thus they will cross paths with each other at most once.

The number of possible Cartesian trajectories in this domain is $N_{c} = |\pazocal{T}|^n(|\pazocal{U}|^n)^H$. Enumerating these trajectories runs in time $O(2^{nH\log\pazocal{U}})$. For the same scene, the number of possible braids generally depends on the structure of the road network. However, we can bound the number of possible outcomes as $N_b \leq 3^{\binom{n}{2}}$, where the exponent is the binomial coefficient representing the number of all pairs of agents and the base represents the 3 types of possible interactions per pair that could be represented by a braid, i.e., ``over-crossing", ``under-crossing", or no crossing. This enumeration runs in time $O(2^{n^2})$.

\theorem\label{theorem:one}
The runtime of enumerating braids is lower than the runtime of enumerating Cartesian trajectories for the class of driving problems considered above.

\proof
We want to show that $2^{n^2} < 2^{nH\log\pazocal{U}}$. This inequality is equivalent to $n < H\log\pazocal{U}$. We assumed that $n\ll H$, $n \ll\pazocal{U}$, therefore it should also hold that $n \ll H\log\pazocal{U}$. Thus the initial inequality holds and supports the statement that the runtime of enumerating braids is significantly lower than the runtime of enumerating Cartesian trajectories. $\blacksquare$

\begin{figure}
\centering
\begin{subfigure}{.48\linewidth}
\centering
\includegraphics[width = \linewidth]{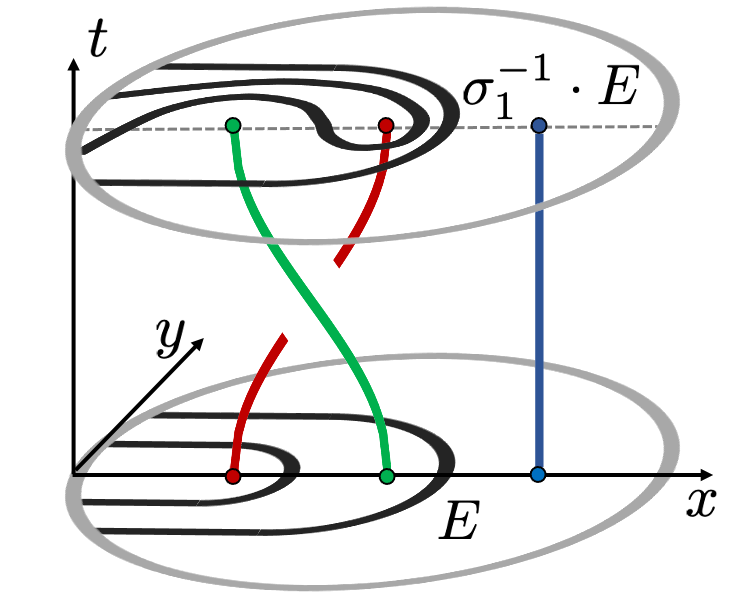}
\caption{Curve diagram for $\sigma_1^{-1}$.\label{fig:curve-diagrams1}}
\end{subfigure}
~
\begin{subfigure}{.48\linewidth}
\centering
\includegraphics[width = \linewidth]{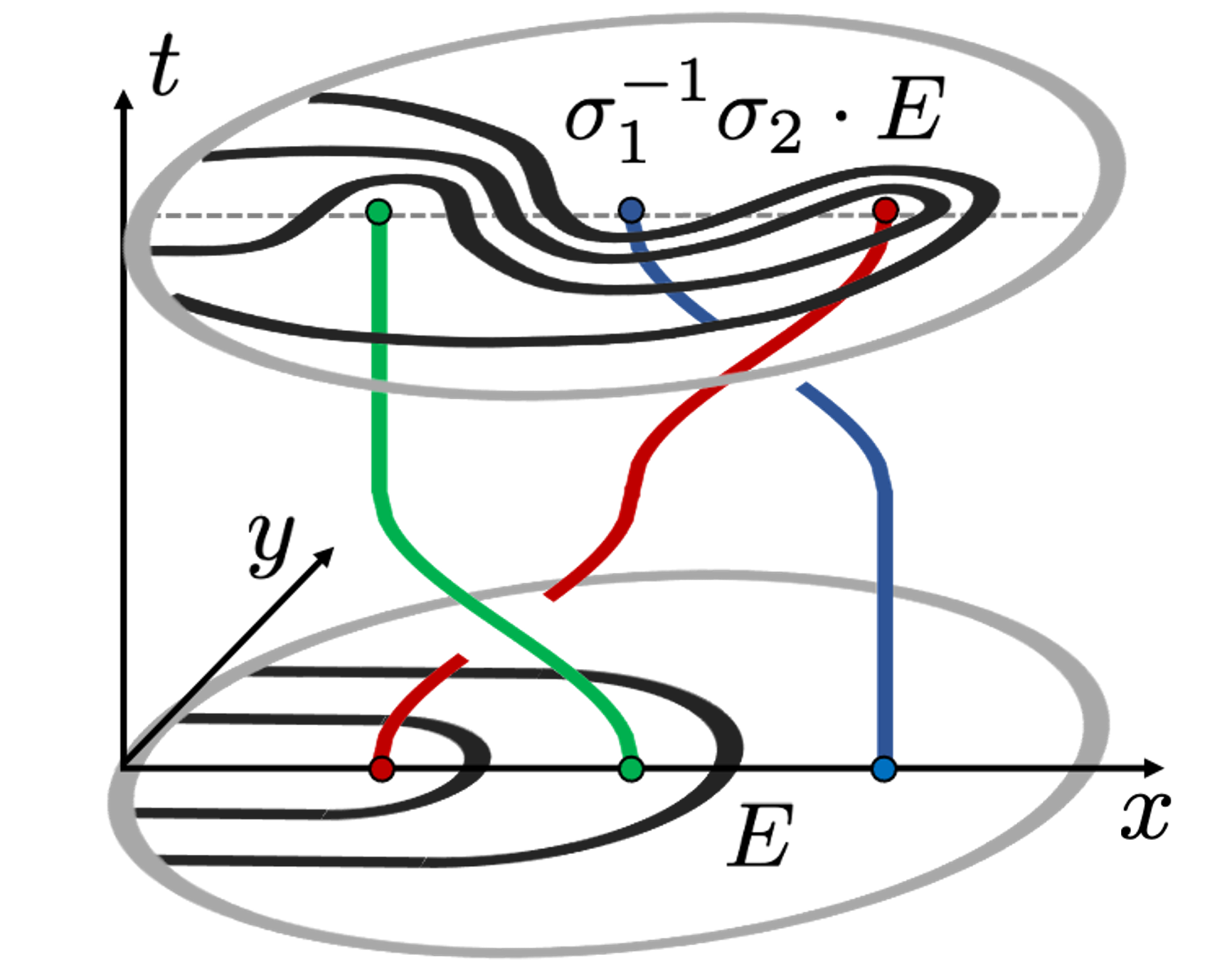}
\caption{Curve diagram for $\sigma_1^{-1}\sigma_{2}$.\label{fig:curve-diagrams2}}
\end{subfigure}
\caption{Curve diagrams for braids of different complexity. The braid $\sigma_{1}^{-1}\sigma_{2}$ (\subref{fig:curve-diagrams2}) is more complex ($TC = 2$) than the braid $\sigma_{1}^{-1}$ ($TC = 1.585$) shown in (\subref{fig:curve-diagrams1}). This is reflected in the higher number of intersections between the curve diagram $\sigma_{1}^{-1}\sigma_{2}\cdot E$ and the $x$-axis (dotted line).\label{fig:example_curve_diagrams}}
\end{figure}

\subsection{Complexity of Braid Entanglement}

The entanglement of the trajectories described by a braid is indicative of the complexity of the interaction between agents. We quantify braid complexity using the \emph{Topological Complexity} index (TC) of \citet{Dynnikov2007} for which we provide an informal definition below.

Assume that a braid $\beta\in B_{n}$ represents the collective motion of $n$ agents from initial locations $\beta(0)$ to final locations $\beta(1)$. Denote by $D^2$ a closed disk surrounding agents' initial positions, $\beta(0)$. Define by $E$ a set of $n-1$ disjoint arcs, anchored on the disk, and separating the agents for $t=0$, defining $n-1$ distinct regions in the disk (see~\figref{fig:example_curve_diagrams}). Assume that these regions are rigidly attached on the agents. As the agents follow the motion described by $\beta$ from $t=0$ to $t=1$, the regions dynamically deform. The image $D=\beta\cdot E$ representing the shape of the regions obtained upon applying the motion described by $\beta$ on $E$ is called a \emph{curve diagram}. The \emph{norm} of curve diagram $D$ is defined as the number of intersections of $D$ with the $x$ axis. 
Based on the above definitions, we can define the TC index of a braid $\beta\in B_{n}$ as:
\begin{equation}TC(\beta) = \log_{2}(||\beta\cdot E||)-\log_{2}(||E||)\mbox{.}\end{equation}
This expression is equivalent to the logarithm of the \emph{gain} of intersections with the $x$-axis, upon application of a braid. \figref{fig:example_curve_diagrams} depicts curve diagrams acquired upon inducing motion of two different braids on the canonical curve diagram $E$.
\section{A Case Study on Traffic Datasets}\label{sec:datasets}

We demonstrate how braids may abstract traffic episodes through a case study on real-world datasets.  

\begin{figure}
\centering
\begin{subfigure}{.23\linewidth}
\centering
\includegraphics[width = \linewidth]{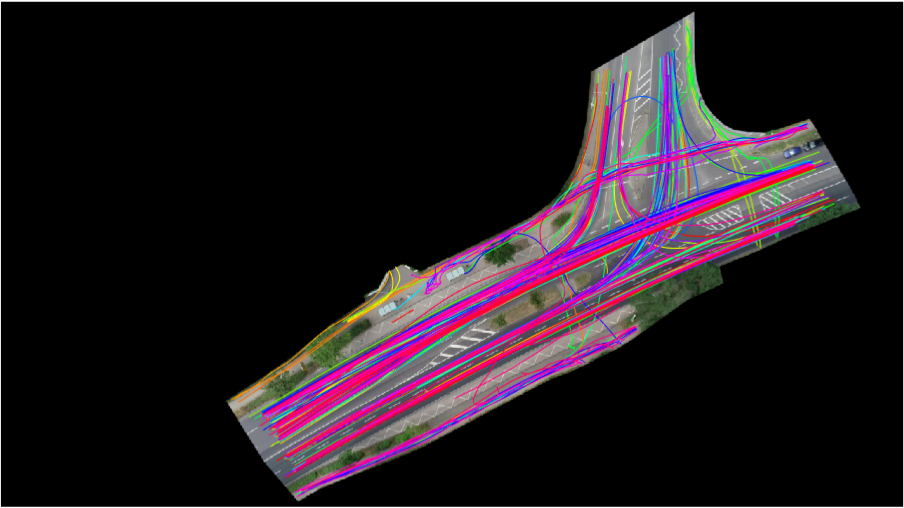}
\caption{inD 1.\label{fig:intersection1}}
\end{subfigure}
\begin{subfigure}{.23\linewidth}
\centering
\includegraphics[width = \linewidth]{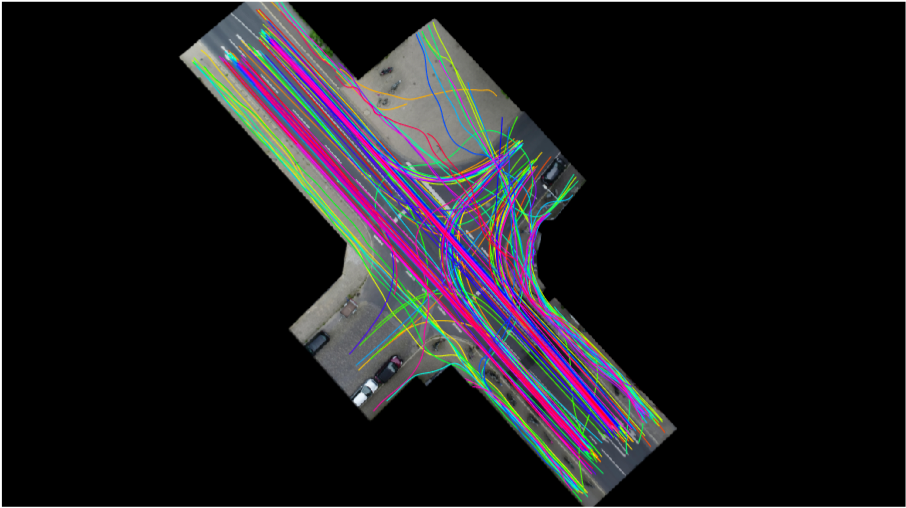}
\caption{inD 2.\label{fig:intersection2}}
\end{subfigure}
\begin{subfigure}{.23\linewidth}
\centering
\includegraphics[width = \linewidth]{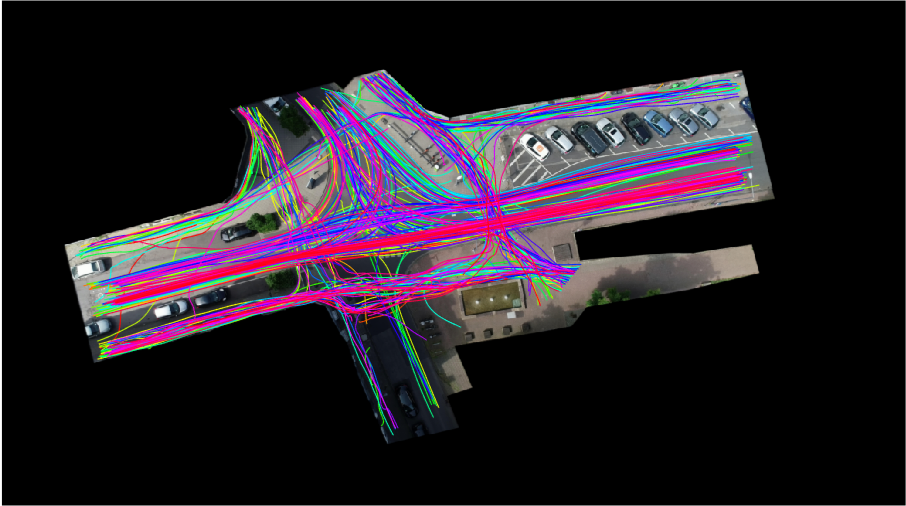}
\caption{inD 3.\label{fig:intersection3}}
\end{subfigure}
\begin{subfigure}{.23\linewidth}
\centering
\includegraphics[width = \linewidth]{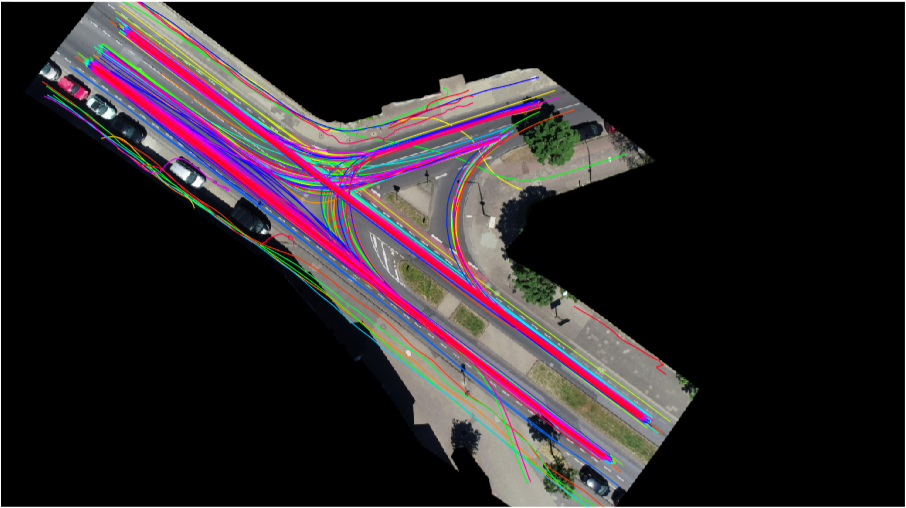}
\caption{inD 4.\label{fig:intersection4}}
\end{subfigure}
\\
\begin{subfigure}{.23\linewidth}
\centering
\includegraphics[width = \linewidth]{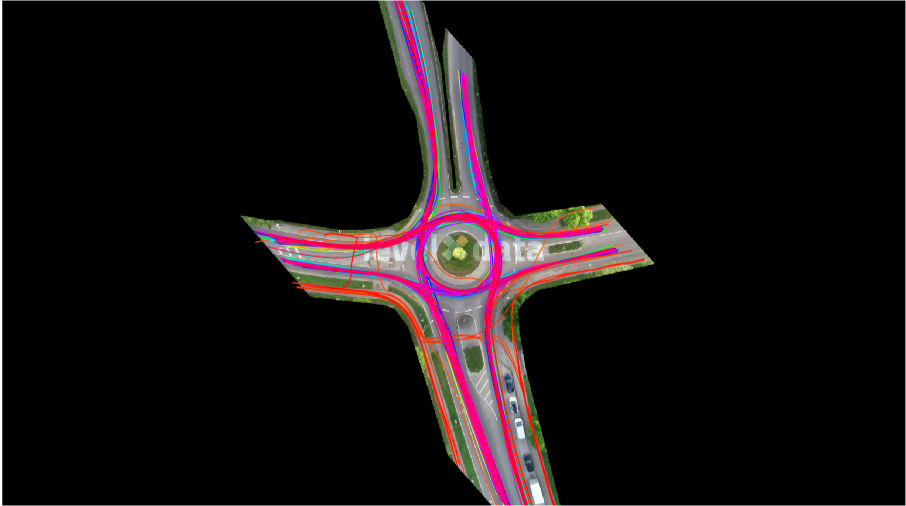}
\caption{rounD 1.\label{fig:roundabouts1}}
\end{subfigure}
\begin{subfigure}{.23\linewidth}
\centering
\includegraphics[width = \linewidth]{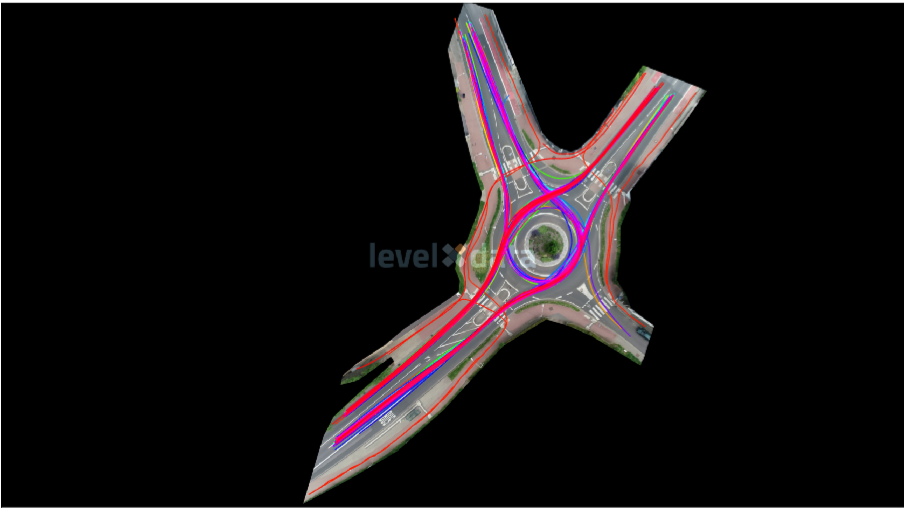}
\caption{rounD 2.\label{fig:roundabouts2}}
\end{subfigure}
\begin{subfigure}{.23\linewidth}
\centering
\includegraphics[width = \linewidth]{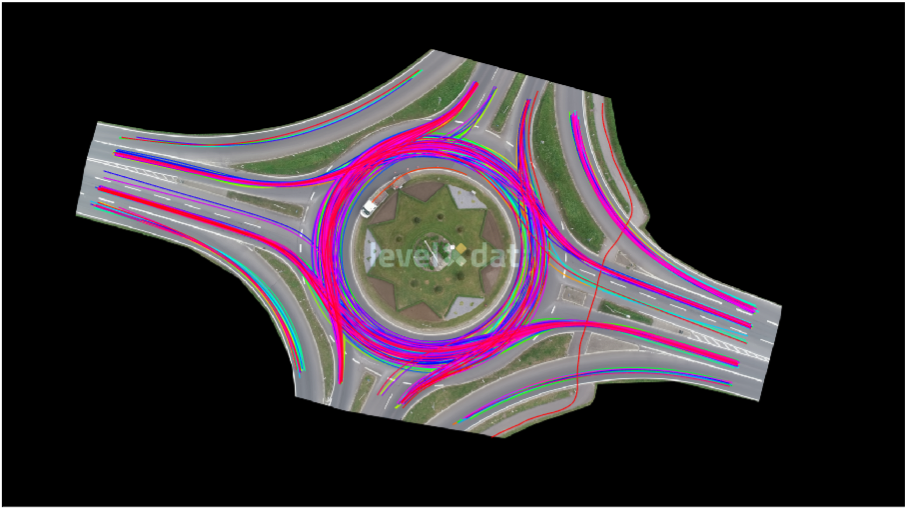}
\caption{rounD 3.\label{fig:roundabouts3}}
\end{subfigure}
\begin{subfigure}{.23\linewidth}
\centering
\includegraphics[width = \linewidth]{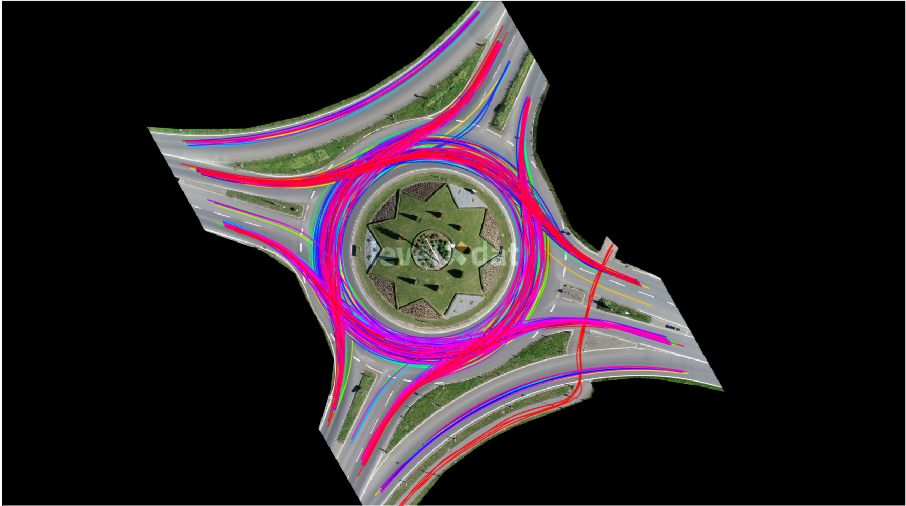}
\caption{rounD 4.\label{fig:roundabouts4}}
\end{subfigure}
\caption{Top view of the 8 scenes from the inD and rounD datasets that we analyzed using topological tools. All trajectories are overlayed on top of the street structures.\label{fig:dataset-scenes}}
\end{figure}

\subsection{Datasets}

We consider the inD~\citep{inDdataset} and rounD~\citep{rounDdataset} datasets, containing trajectory data (of vehicles, pedestrians and bicycles) recorded respectively in four intersection scenes and four roundabout scenes of the German road network. Both datasets were extracted from drone footage in $\SI{25}{\fps}$ via computer vision techniques, yielding an estimated positional error in the order of $\SI{10}{\cm}$. A top view of the eight scenes is shown in~\figref{fig:dataset-scenes}, and their dimensions are listed in Table~\ref{tab:scene-details}.

\subsection{Methodology}

We split each scene into a set of sequential episodes, sweeping the whole duration of the recording. Each episode has a fixed duration of $\Delta T = \SI{10}{\second}$, containing trajectories of simultaneously navigating agents. From qualitative inspection, we observed that the most interesting interactions across all scenes involved vehicle traffic; to highlight dynamic vehicle traffic, we filtered out agents moving with a speed lower than $\SI{14}{\metre\per\second}$ and agents that are too far from each other (agents that kept a minimum distance greater than $d_{min} = \SI{10}{\metre}$ throughout the episode). This resulted in a set of episodes summarized in Table~\ref{tab:scene-details}. Using the framework of Sec.~\ref{sec:braids}, we abstracted the trajectory of each episode into a braid, leveraging the braid relations of eq.~\eqref{eq:braidrelations}. Finally, we computed the TC score for each braid. We performed all computations using the Braidlab package~\citep{braidlab}.



\begin{figure}
\centering
\begin{subfigure}{.48\linewidth}
\centering
\includegraphics[trim={11cm 1cm 14.5cm 4cm},clip,width = \linewidth]{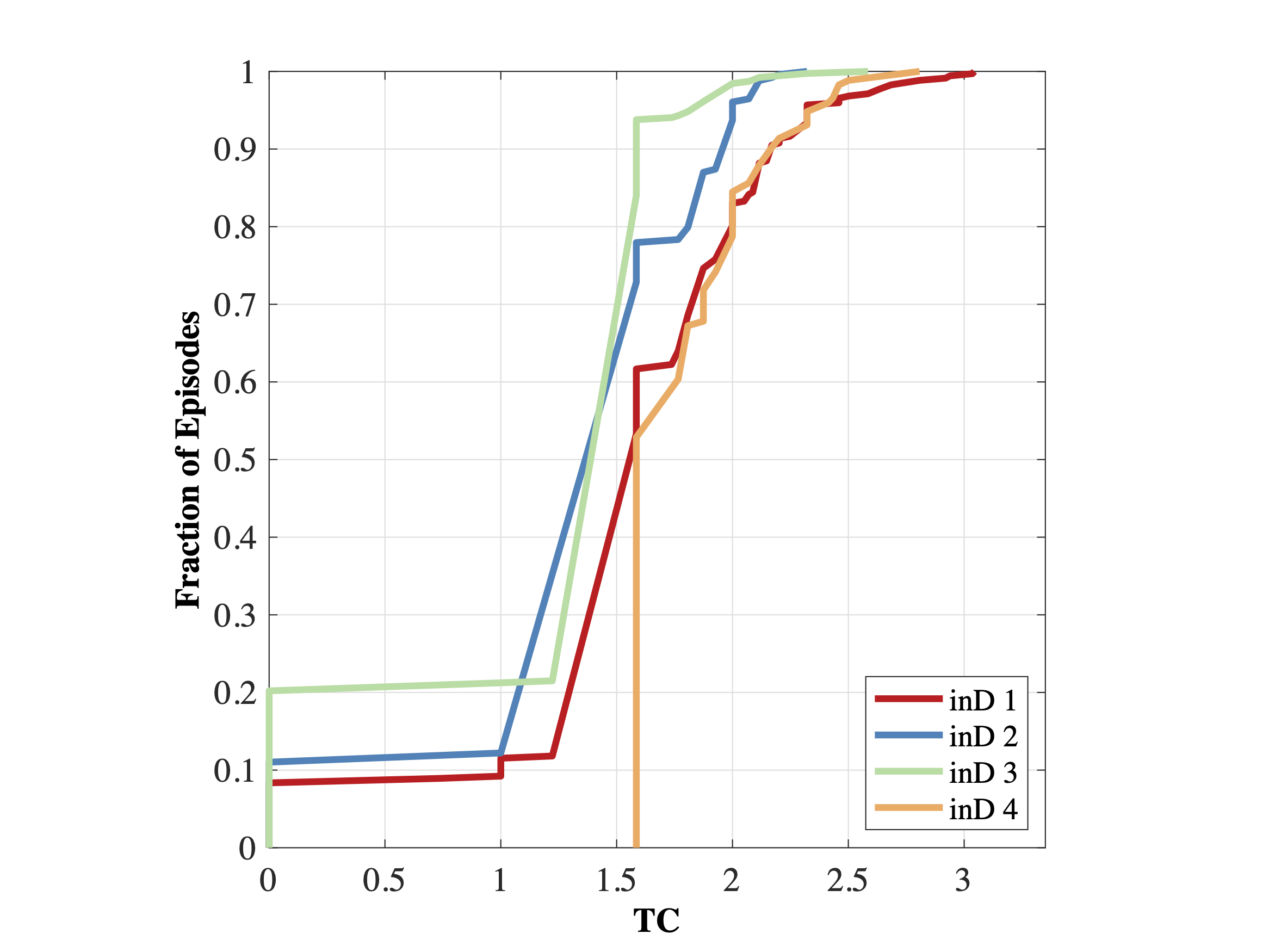}
\caption{inD scenes.\label{fig:intersections-cumulative}}
\end{subfigure}
~
\begin{subfigure}{.48\linewidth}
\centering
\includegraphics[trim={11cm 1cm 14.5cm 4cm},clip,width = \linewidth]{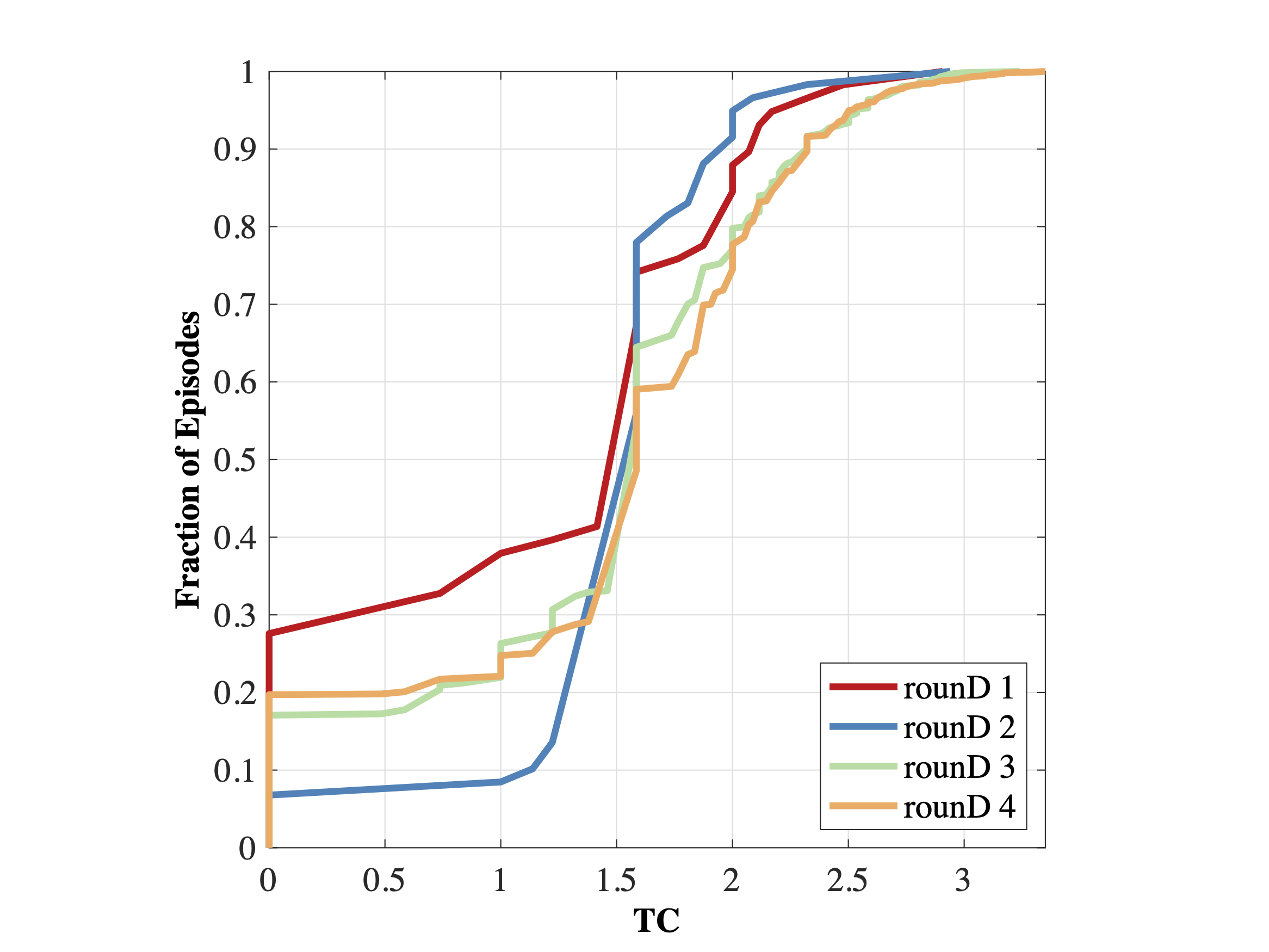}
\caption{rounD scenes.\label{fig:roundabouts-cumulative}}
\end{subfigure}
\caption{Cumulative density of TC (Topological Complexity index) in intersections (\subref{fig:intersections-cumulative}) and roundabouts (\subref{fig:roundabouts-cumulative}).\label{fig:cumulative-plots}}
\end{figure}


\begin{figure}[t]
\centering
\begin{subfigure}{.48\linewidth}
\centering
\includegraphics[width = \linewidth]{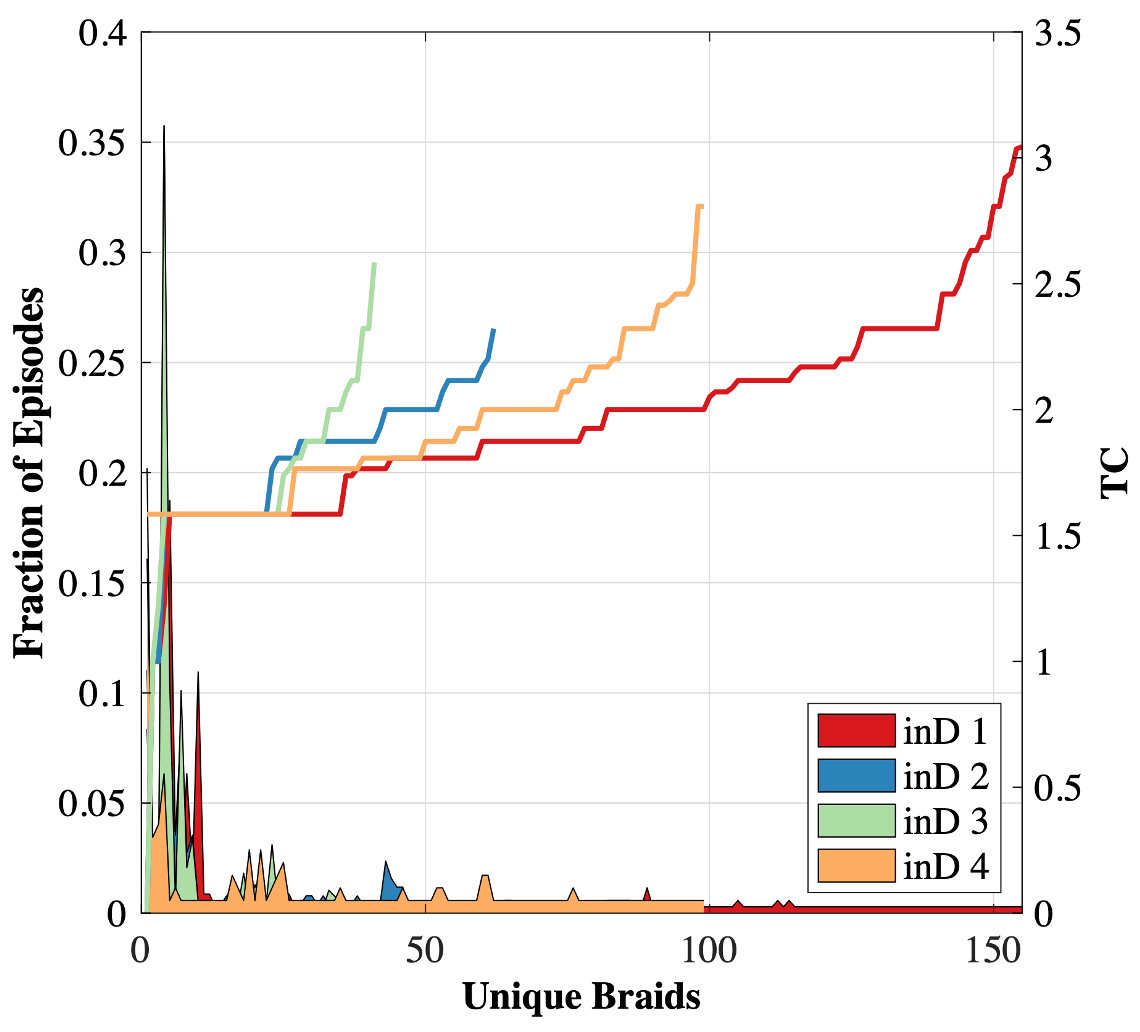}
\caption{inD scenes.\label{fig:intersections-counts-TC}}
\end{subfigure}
~
\begin{subfigure}{.48\linewidth}
\centering
\includegraphics[width = \linewidth]{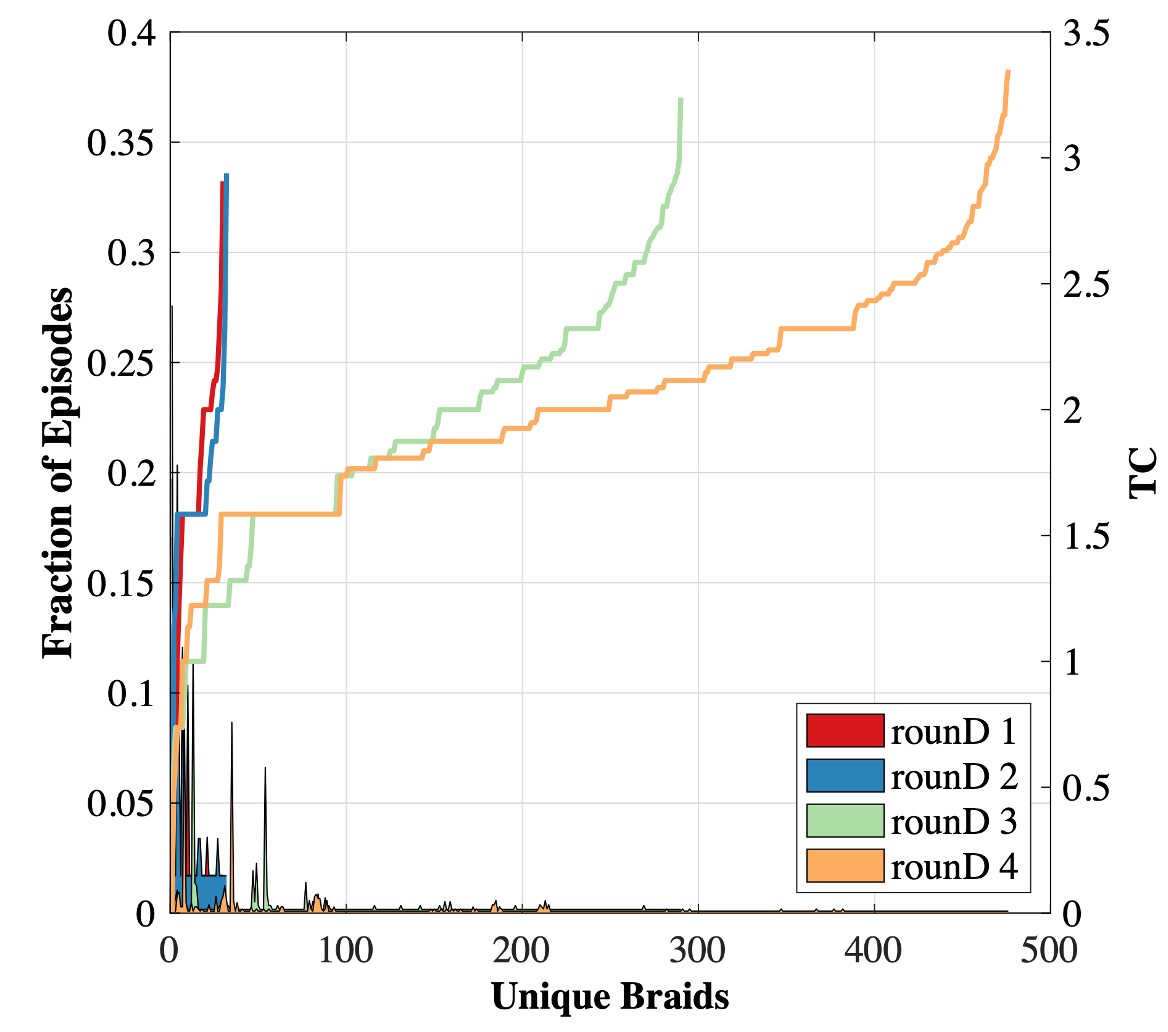}
\caption{rounD scenes.\label{fig:roundabouts-counts-TC}}
\end{subfigure}
\caption{Frequency of unique braids in intersections (\subref{fig:intersections-counts-TC}) and roundabouts (\subref{fig:roundabouts-counts-TC}), arranged in order of increasing TC (Topological Complexity index).\label{fig:cumulative-plots-counts}}
\end{figure}

\subsection{Analysis}

\begin{figure*}
\centering
\begin{subfigure}{.31\linewidth}
\centering
\includegraphics[width = \linewidth]{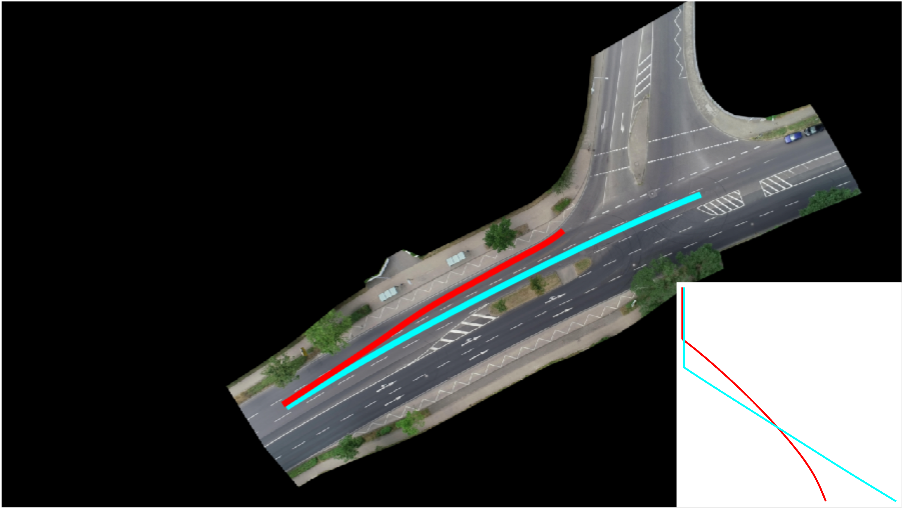}
\caption{inD 1, $TC = 0$.\label{fig:simpleintersection1}}
\end{subfigure}
\begin{subfigure}{.31\linewidth}
\centering
\includegraphics[width = \linewidth]{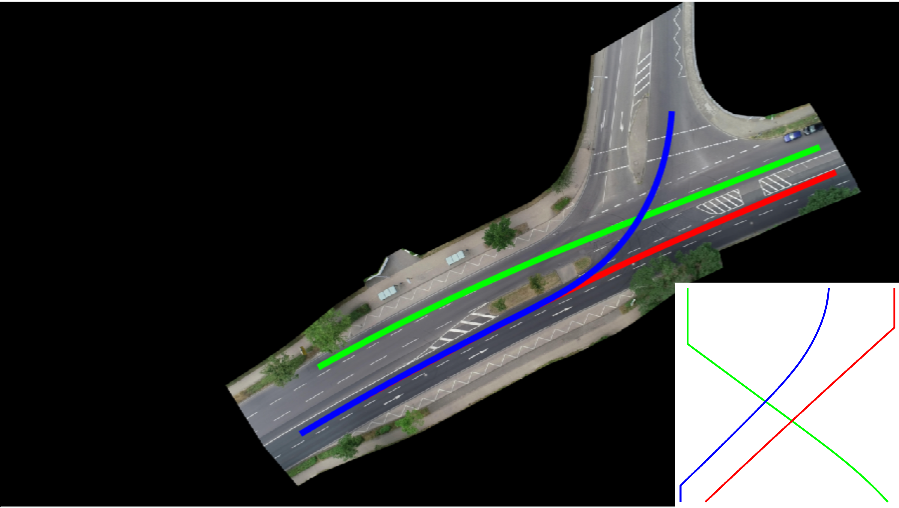}
\caption{inD 1, $TC = 1.5850$.\label{fig:midcomplexintersection1}}
\end{subfigure}
\begin{subfigure}{.31\linewidth}
\centering
\includegraphics[width = \linewidth]{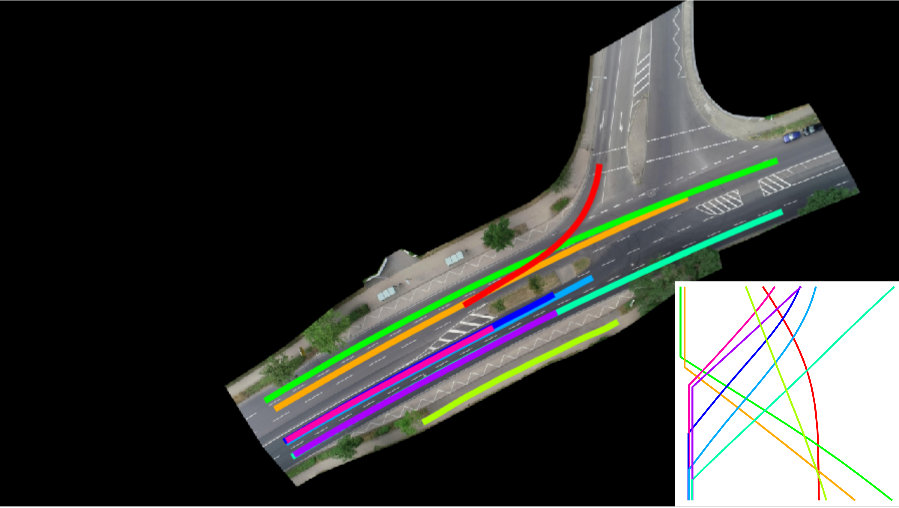}
\caption{inD 1, $TC = 3.0444$.\label{fig:complexintersection1}}
\end{subfigure}
\\
\begin{subfigure}{.31\linewidth}
\centering
\includegraphics[width = \linewidth]{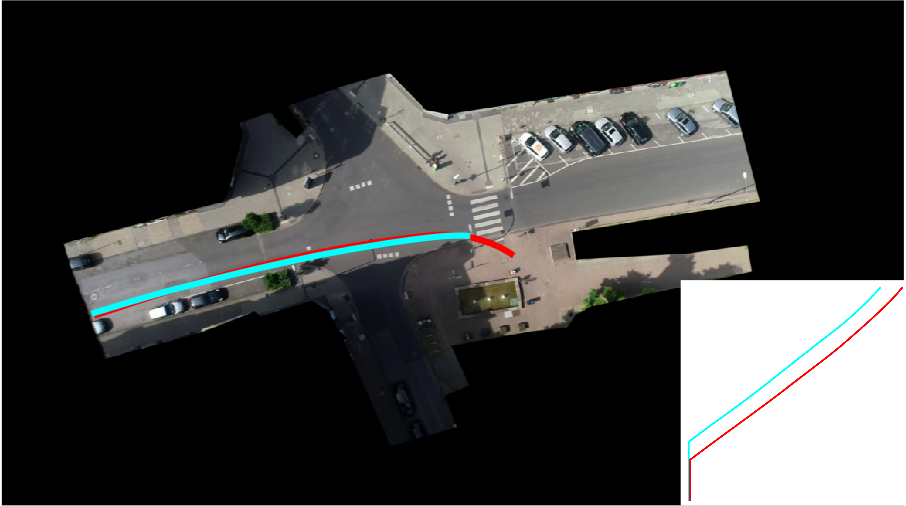}
\caption{inD 3, $TC = 0$.\label{fig:simpleintersection2}}
\end{subfigure}
\begin{subfigure}{.31\linewidth}
\centering
\includegraphics[width = \linewidth]{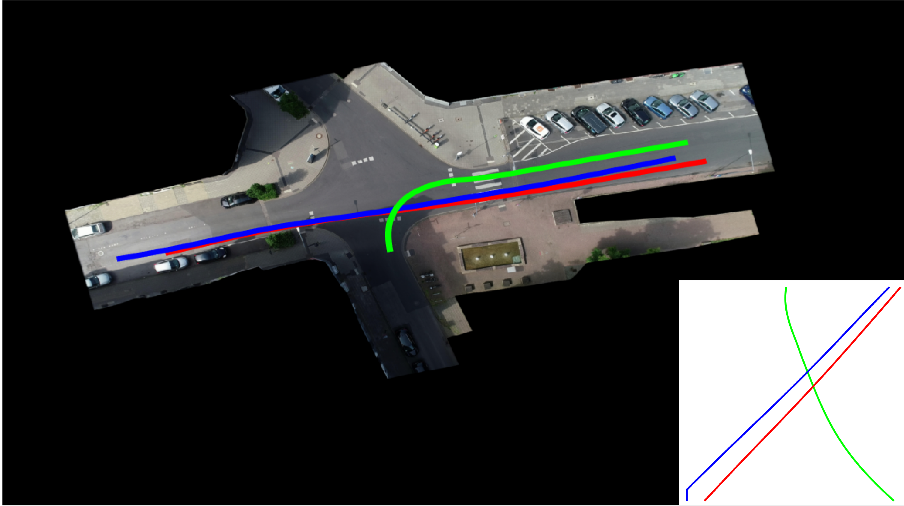}
\caption{inD 3, $TC = 1.5850$.\label{fig:midcomplexintersection2}}
\end{subfigure}
\begin{subfigure}{.31\linewidth}
\centering
\includegraphics[width = \linewidth]{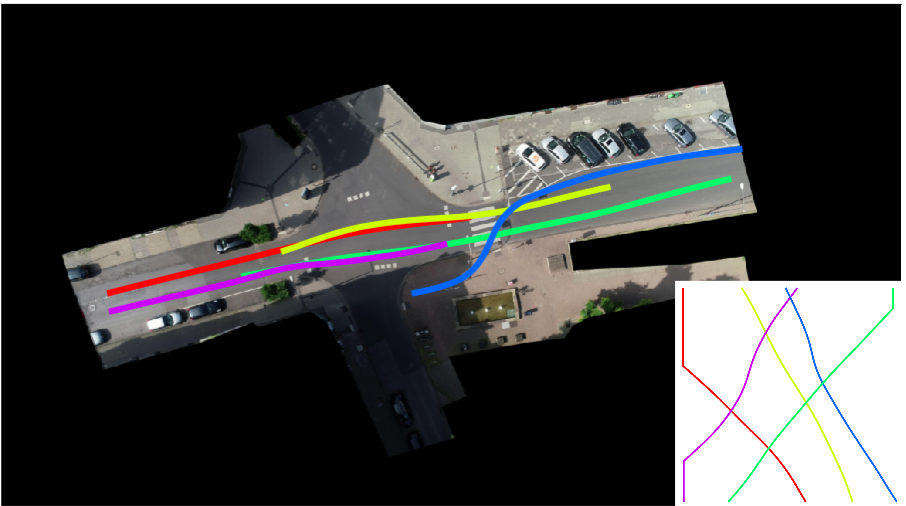}
\caption{inD 3, $TC = 2.5850$.\label{fig:complexintersection2}}
\end{subfigure}
\\
\begin{subfigure}{.31\linewidth}
\centering
\includegraphics[width = \linewidth]{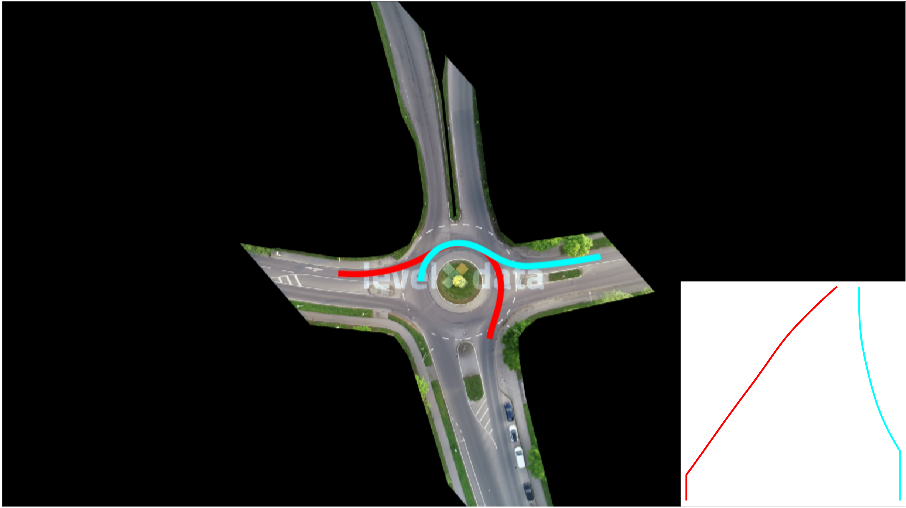}
\caption{rounD 1, $TC = 0$.\label{fig:simpleroundabout3}}
\end{subfigure}
\begin{subfigure}{.31\linewidth}
\centering
\includegraphics[width = \linewidth]{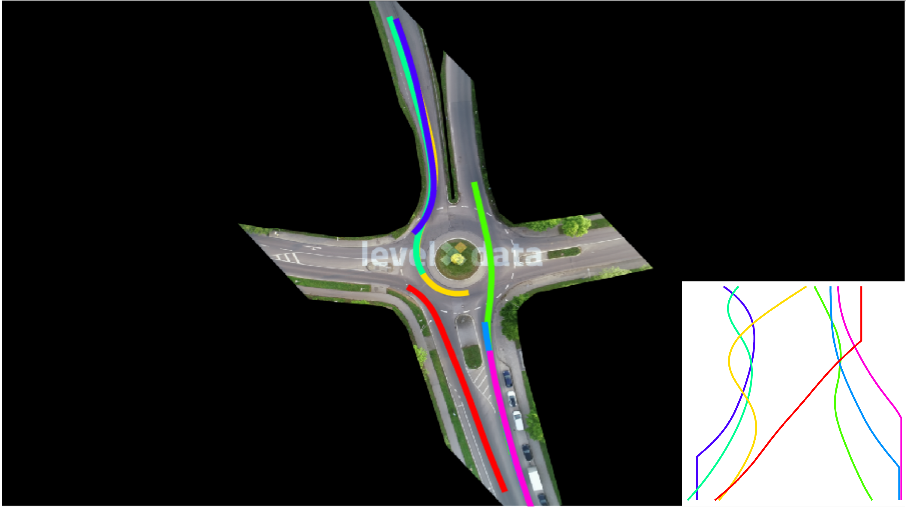}
\caption{rounD 1, $TC =1.4150$.\label{fig:midcomplexroundabout3}}
\end{subfigure}
\begin{subfigure}{.31\linewidth}
\centering
\includegraphics[width = \linewidth]{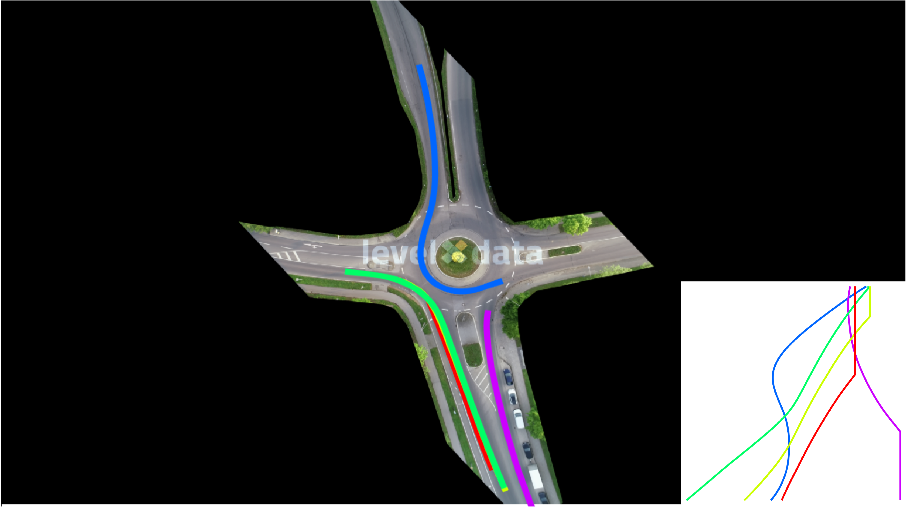}
\caption{rounD 1, $TC = 2.9069$.\label{fig:complexroundabout3}}
\end{subfigure}
\\
\begin{subfigure}{.31\linewidth}
\centering
\includegraphics[width = \linewidth]{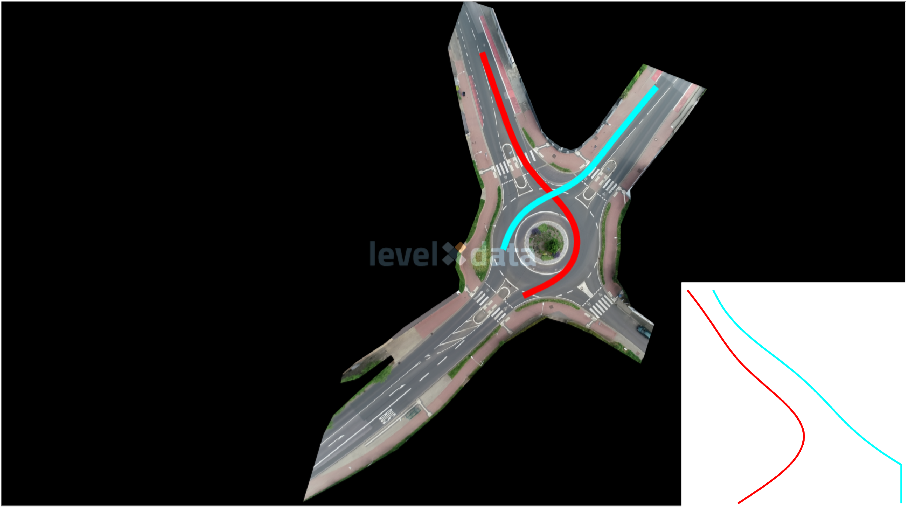}
\caption{rounD 2, $TC = 0$.\label{fig:simpleroundabout4}}
\end{subfigure}
\begin{subfigure}{.31\linewidth}
\centering
\includegraphics[width = \linewidth]{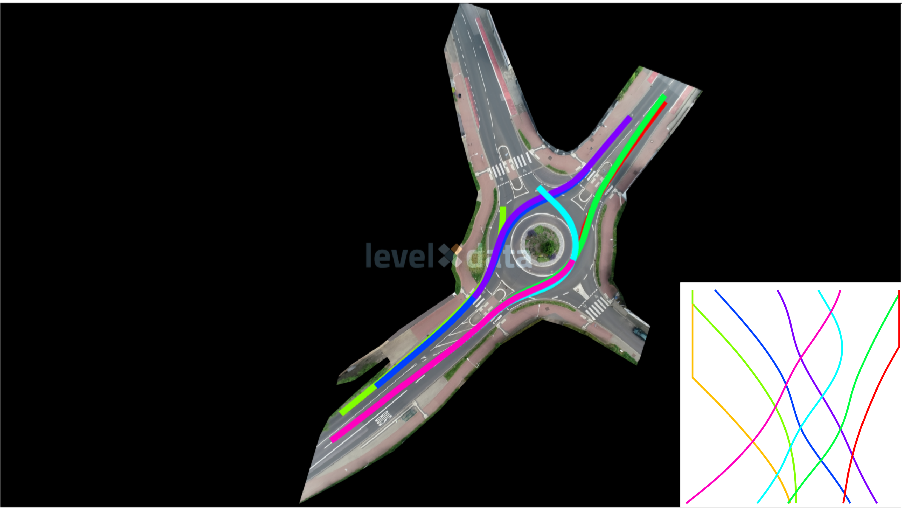}
\caption{rounD 2, $TC = 1.7162$.\label{fig:midcomplexroundabout4}}
\end{subfigure}
\begin{subfigure}{.31\linewidth}
\centering
\includegraphics[width = \linewidth]{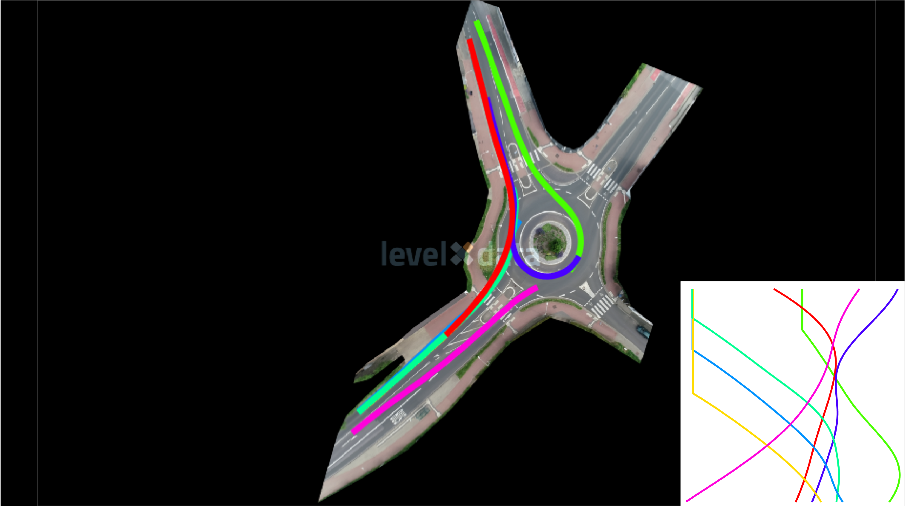}
\caption{rounD 2, $TC = 2.9386$.\label{fig:complexroundabout4}}
\end{subfigure}
\\
\begin{subfigure}{.31\linewidth}
\centering
\includegraphics[width = \linewidth]{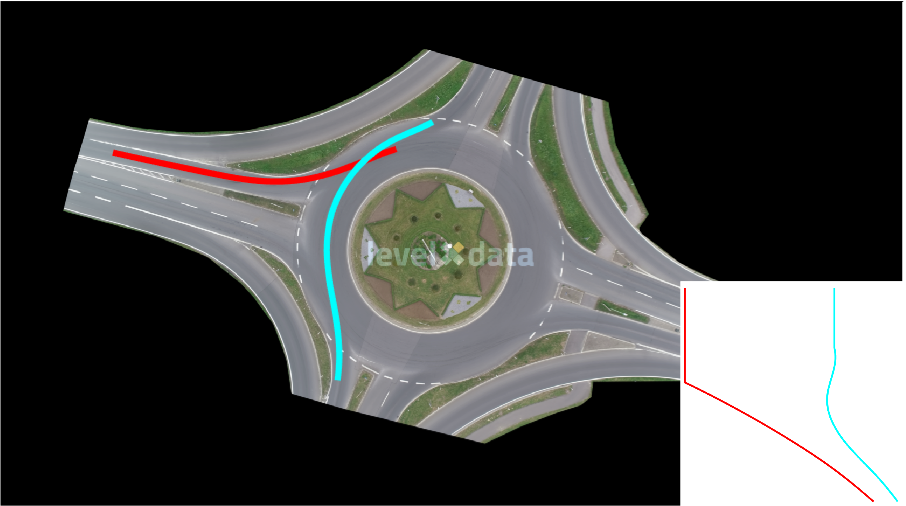}
\caption{rounD 3, $TC = 0$.\label{fig:simpleroundabout5}}
\end{subfigure}
\begin{subfigure}{.31\linewidth}
\centering
\includegraphics[width = \linewidth]{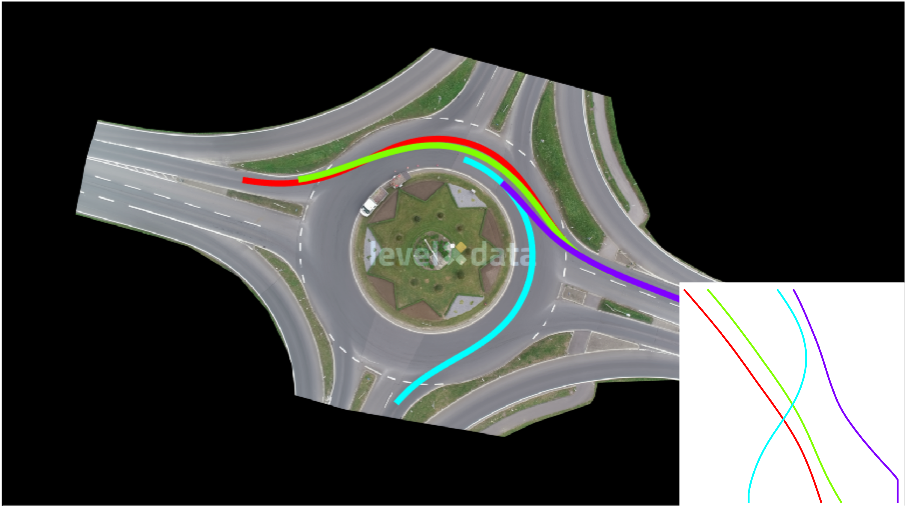}
\caption{rounD 3, $TC = 1.2224$.\label{fig:midcomplexroundabout5}}
\end{subfigure}
\begin{subfigure}{.31\linewidth}
\centering
\includegraphics[width = \linewidth]{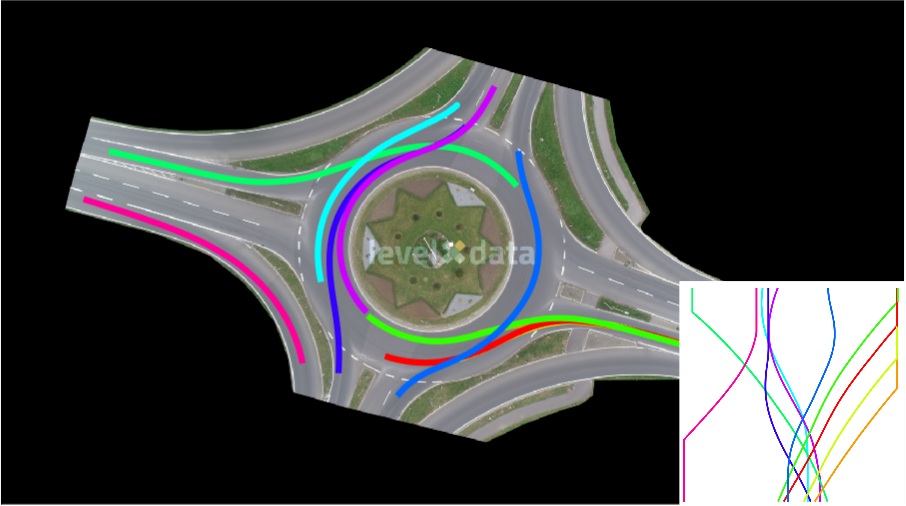}
\caption{rounD 3, $TC = 3.2395$.\label{fig:complexroundabout5}}
\end{subfigure}
\\
\begin{subfigure}{.31\linewidth}
\centering
\includegraphics[width = \linewidth]{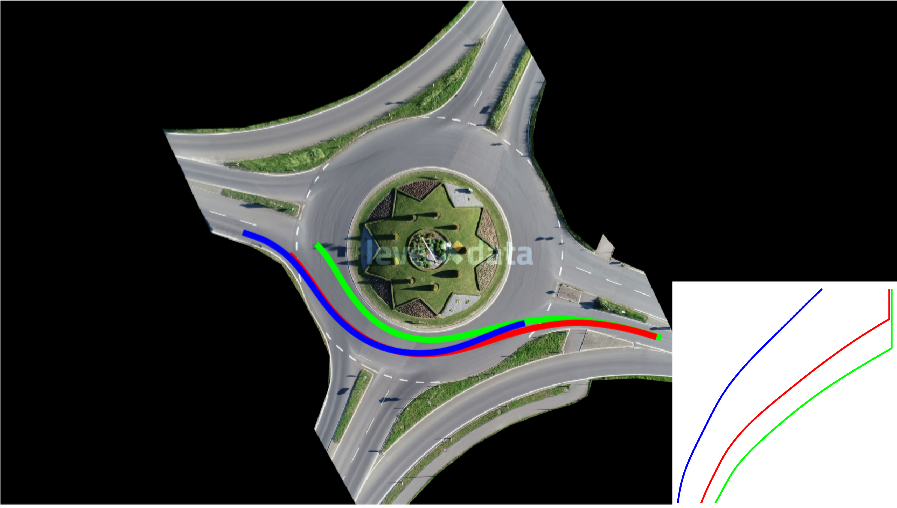}
\caption{rounD 4, $TC = 0$.\label{fig:simpleroundabout6}}
\end{subfigure}
\begin{subfigure}{.31\linewidth}
\centering
\includegraphics[width = \linewidth]{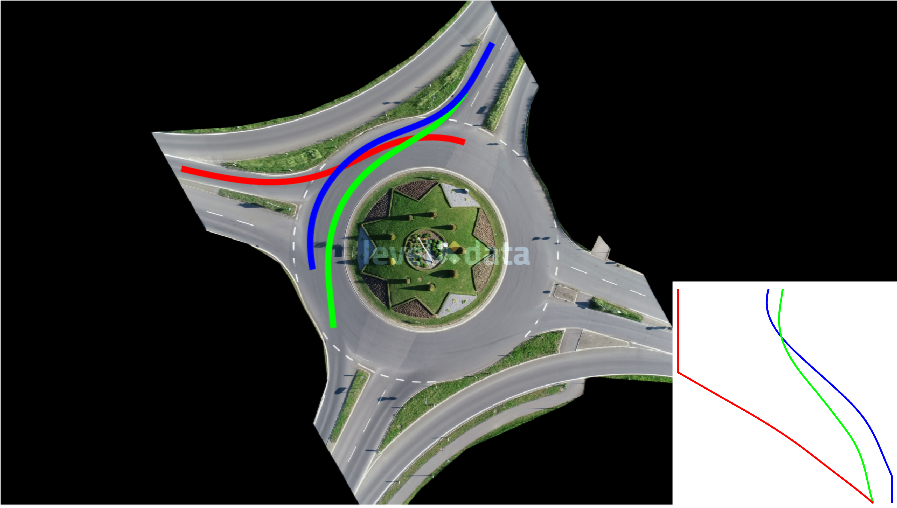}
\caption{rounD 4, $TC = 1$.\label{fig:midcomplexroundabout6}}
\end{subfigure}
\begin{subfigure}{.31\linewidth}
\centering
\includegraphics[width = \linewidth]{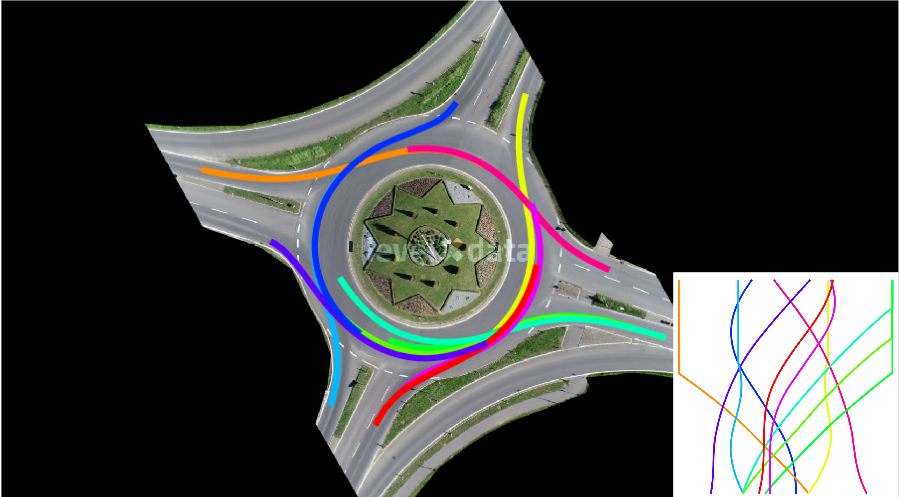}
\caption{rounD 4, $TC = 3.3505$.\label{fig:complexroundabout6}}
\end{subfigure}
\caption{Episodes with different Topological Complexity (TC). Each row depicts three episodes yielding distinct braids in the same scene. At the bottom right of each figure, the braid formed by the data through a $x$-$t$ side projection of the episode is plotted. The episodes on each row are organized from left to the right in order of increasing TC. In all scenes, the agents are following the right-hand traffic convention. \label{fig:TC-examples}}
\end{figure*}

The behavior in each scene is clustered into a small number of unique braids, describing vehicles' interaction patterns (see Table~\ref{tab:scene-details}). This highlights that real-world traffic tends to collapse to a small set of outcomes. The extracted braids are mapped onto the TC values on the right. \figref{fig:TC-examples} depicts episodes of varying TC, drawn from the two datasets, along with their braid representatives and TC scores. We see that complex interactions get mapped onto higher TC values.

\figref{fig:cumulative-plots} shows the empirical cumulative density of TC across the inD and rounD dataset scenes. We see that each scene has a distinct complexity growth pattern but in both datasets, about $60\%$ of episodes are concentrated below $TC = 1.5$. This is highlighted in~\figref{fig:cumulative-plots-counts}, which shows the relative frequency of unique braids per scene, organized in order of increasing TC. We see that the mass of the frequency is concentrated on the left side for both plots, suggesting that the majority of episodes feature a relatively low degree of interaction. This indicates that despite the dense traffic exhibited in the datasets (Table~\ref{tab:scene-details}), the vast majority of episodes involve traffic that is orderly and well organized. This is an artifact of the underlying spatiotemporal structure (geometry, traffic rules, driving styles).


\subsection{Discussion}



Our representation enables enumeration of the classes of multiagent interaction patterns that are \emph{theoretically} possible in a traffic scene in a compact and interpretable form. Further, given a traffic dataset, it allows us to extract the subset of interaction patterns that are \emph{empirically} relevant. This may inform algorithmic design, benchmarking and even road network design. Importantly, our framework can be valuable for characterizing a traffic dataset with respect to the support it provides over the space of theoretically possible behavior in a domain. Understanding the support of a dataset is crucial for data-driven approaches~\citep{Roh2020Multimodal,trajectron,bgap}  but also for guiding the process of synthetically generating simulated scenarios to produce diverse datasets.





Our framework is complementary to alternative approaches for characterizing interaction, such as the interactivity score~\citep{tolstaya2021identifying} and distribution-based KL-divergence. The Interactivity score may miss crucial interaction events: scores can be large when there is high correlation between two trajectories (e.g., one car following another), but small when trajectories are dissimilar (e.g., cars crossing an intersection). In contrast, TC will account for these situations through the consideration of the underlying topological structure. Further, our framework may be directly applicable to any traffic dataset~\citep{EttingerWaymoDataset,nuscenes2019,argoverse2019} and even to alternative domains like pedestrian tracking~\citep{PellegriniESG09} or sports analysis~\citep{sportvu} without additional modifications. It may complement temporal logic approaches for trajectory labeling~\citep{puranic2021learning, LiRGVDKR21} which often require involved and domain-specific mathematical treatment~\citep{SchulzHLWB17}. 


\begin{table}
\centering
    \caption{Scene details and interaction statistics.}
    \label{tab:scene-details}
\resizebox{\columnwidth}{!}{%
    \begin{tabular}{c|c|c|c|c|c}
        \toprule
         Scene & Dimensions ($\SI{}{\metre}^2$) & Episodes & Agents/Episode (M, SD) & Unique braids & TC (M, SD)\\
        \midrule
         inD 1 & $131\times 110$ & 347 & $3.62\pm 1.76$ & 155 & $1.62 \pm 0.59$\\
         inD 2 & $59\times 64$ & 254 & $2.82\pm 1.00$ & 62 & $1.48 \pm 0.55$\\
         inD 3 & $85\times 45$ & 386 & $2.62\pm 0.90$ & 41 & $1.28 \pm 0.66$\\
         inD 4 & $79\times 67$ & 174 & $4.10\pm 1.51$ & 99 & $1.79 \pm 0.28$\\
         rounD 1 & $99\times 143$ & 58 & $3.16\pm 1.45$ & 30 & $1.20 \pm 0.84$\\
         rounD 2 & $99\times 122$ & 59 & $3.85\pm 1.75$ & 32 & $1.54 \pm 0.50$\\
         rounD 3 & $127\times 69$ & 574 & $4.36\pm 2.28$ & 290 & $1.43 \pm 0.79$\\
         rounD 4 & $92\times 98$ & 1050 & $4.07\pm 2.00$ & 476 & $1.46 \pm 0.83$\\
        \bottomrule
    \end{tabular}
    }
\end{table}

\section{Conclusion}\label{sec:discussion}

Abstractions like braids highlight topological patterns of interaction like vehicles' crossings or overtaking maneuvers through projection transforms or simplification rules like eq.~\ref{eq:braidrelations}. They filter out geometric features, like the temporal spacing between vehicles or a driver's erratic maneuvers. These artifacts could be relevant to traffic analysis. Thus, the proposed framework is not meant to replace existing, geometry-focused tools but rather to complement them. 

Our goal in this study was to demonstrate that tools from braid theory can be valuable for the analysis of multiagent behavior in traffic scenes. Thus, some of the parameters chosen, e.g., the braid projection plane, the episode duration, agents' distance and speed thresholds were not optimally selected. These parameters could be adapted to the specific context of a scene or the scope of an investigation.

\balance
\bibliographystyle{abbrvnat}
\bibliography{references}

\end{document}